\documentclass[lettersize,journal]{IEEEtran}
\usepackage{amsmath,amsfonts}
\usepackage{algorithmic}
\usepackage{algorithm}
\usepackage{array}
\usepackage[caption=false,font=normalsize,labelfont=sf,textfont=sf]{subfig}
\usepackage{textcomp}
\usepackage{stfloats}
\usepackage{url}
\usepackage{verbatim}
\usepackage{graphicx}
\usepackage{cite}
\usepackage{adjustbox}
\usepackage{wrapfig}
\usepackage{xspace}
\usepackage{amssymb}
\usepackage{pifont}
\usepackage{amsmath}
\usepackage{multirow}
\usepackage{booktabs}
\usepackage[colorlinks,linkcolor=red]{hyperref}
\newcommand{\cmark}{\ding{51}}%
\newcommand{\xmark}{\ding{55}}%
\title{TCJA-SNN: Temporal-Channel Joint Attention for Spiking Neural Networks}

\author{
Rui-Jie Zhu,~\IEEEmembership{Graduate Student Member,~IEEE,}
        Malu Zhang,~\IEEEmembership{Member,~IEEE,}
        Qihang Zhao,\\
        Haoyu Deng,
        Yule Duan,
        and Liang-Jian Deng,~\IEEEmembership{Senior Member,~IEEE}
        
\thanks{R. J. Zhu is with Department of Electrical Computer Engineering, University of California, Santa Cruz, CA, USA and School of Public Affairs and Administration, University of Electronic Science and Technology of China, Chengdu, China}
\thanks{M. Zhang, H. Deng, Y. Duan, and L. J. Deng are with the University of Electronic Science and Technology of China.}
\thanks{Q. Zhao is with the Kuaishou Technology, Beijing, China}
\thanks{Corresponding author: Malu Zhang and Liang-Jian Deng.}%

}

\usepackage{bibentry}

\begin{document}

\maketitle

\begin{abstract}

Spiking Neural Networks (SNNs) are attracting widespread interest due to their biological plausibility, energy efficiency, and powerful spatio-temporal information representation ability. Given the critical role of attention mechanisms in enhancing neural network performance, the integration of SNNs and attention mechanisms exhibits tremendous potential to deliver energy-efficient and high-performance computing paradigms.  In this paper, we present a novel Temporal-Channel Joint Attention mechanism for SNNs, referred to as TCJA-SNN. The proposed TCJA-SNN framework can effectively assess the significance of spike sequence from both spatial and temporal dimensions. More specifically,  our essential technical contribution lies on:
1) We employ the squeeze operation to compress the spike stream into an average matrix. Then, we leverage two local attention mechanisms based on efficient 1-D convolutions to facilitate comprehensive feature extraction at the temporal and channel levels independently. 2) We introduce the Cross Convolutional Fusion (CCF) layer as a novel approach to model the inter-dependencies between the temporal and channel scopes. This layer effectively breaks the independence of these two dimensions and enables the interaction between features. Experimental results demonstrate that the proposed TCJA-SNN outperforms the state-of-the-art on all standard static and neuromorphic datasets, including Fashion-MNIST, CIFAR10, CIFAR100, CIFAR10-DVS, N-Caltech 101, and DVS128 Gesture.
Furthermore, we effectively apply the TCJA-SNN framework to image generation tasks by leveraging a variation autoencoder. To the best of our knowledge, this study is the first instance where the SNN-attention mechanism has been employed for high-level classification and low-level generation tasks. \textit{Our implementation codes are available at \url{https://github.com/ridgerchu/TCJA}.}

\end{abstract}
\raggedbottom
\section{Introduction}
Spiking Neural Networks (SNNs) have emerged as a promising research area, offering lower energy consumption and superior robustness compared to conventional Artificial Neural Networks (ANNs) \cite{roy2019towards, stromatias2015robustness}. These characteristics make SNNs highly promising for temporal data processing and power-critical applications~\cite{roy2019towards,zhang2021rectified}. {In recent years, significant progress has been made by incorporating backpropagation into SNNs \cite{bohte2000spikeprop,wu2021progressive,dampfhoffer2023backpropagation,lee2016training, luo2022supervised,zhang2021rectified, cao2015spiking, zhang2023self},} which allows the integration of various ANN modules into SNN architectures, including batch normalization blocks \cite{zheng2021going} and residual blocks \cite{hu2018spiking}. By leveraging these ANN-based methods, it becomes possible to train large-scale SNNs while preserving the energy efficiency associated with SNN's binary spiking nature. 

Despite significant progress, SNNs have yet to fully exploit the superior representational capability of deep learning, primarily due to their unique training mode, which struggles to model complex channel-temporal relationships effectively. {To address this limitation, Zheng \textit{et al.}~\cite{zheng2021going} introduced a batch normalization method for the temporal dimension, overcoming issues of gradient vanishing and threshold-input balance. On the other hand, Wu \textit{et al.}~\cite{wu2019direct} proposed a method named NeuNorm to address the channel-wise challenges. }NeuNorm includes an auxiliary neuron that adjusts the stimulus strength generated by the preceding layer, enhancing performance while mimicking the activity of the retina and nearby cells for added bio-plausibility. However, existing methods handle temporal and channel information separately, leading to limited joint information extraction. Given that SNNs reuse network parameters at each time step, there exists untapped potential for recalibration at both the temporal and channel dimensions. Especially, TA-SNN proposed by Yao et al.~\cite{yao2021temporal}.

\begin{figure*}
    \centering
    \includegraphics[width=0.60\linewidth]{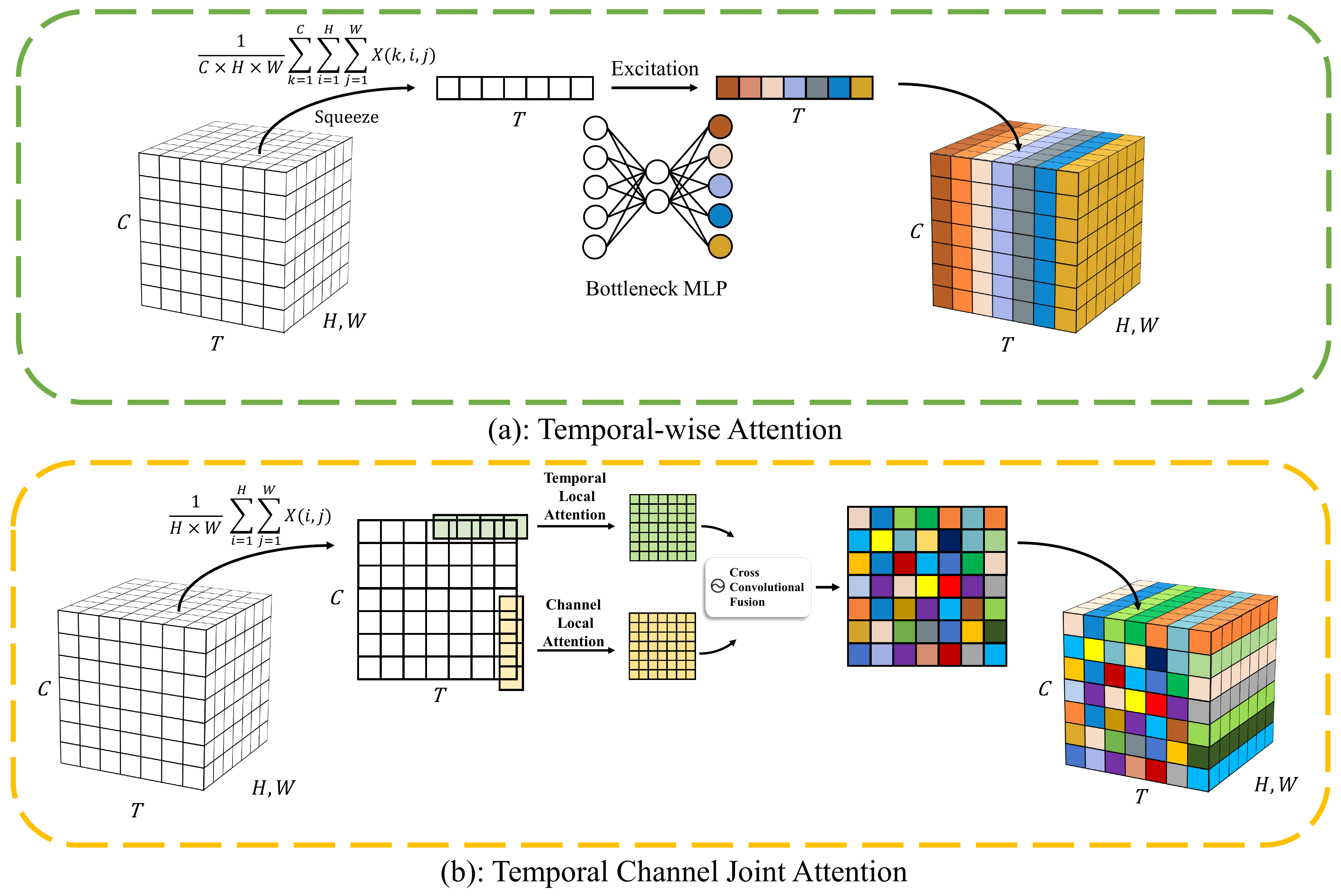}
    \caption{How our Temporal-Channel Joint Attention differs from existing temporal-wise attention~\cite{yao2021temporal}, which estimates the saliency of each time step by squeeze-and-excitation module. $T$ denotes the time step, $C$ denotes the channel, and $H,W$ represents the spatial resolution. By utilizing two separate 1-D convolutional layers and the Cross Convolutional Fusion (CCF) operation, our Temporal-Channel Joint Attention establishes the association between the time step and the channel.}
    \label{fig_general}
\end{figure*}

Previous studies in ANNs~\cite{hu2018squeeze, woo2018cbam} have often utilized the attention mechanism as a means to address the challenges posed by multidimensional dynamic problems. The attention mechanism, inspired by human cognitive processes, enables the selective focus on relevant information while disregarding irrelevant data. This approach has shown promise in the realm of SNNs and merits further exploration~\cite{yao2021temporal}. {For instance, in the domain of neuroscience, an attention-based Spike-Timing-Dependent Plasticity (STDP) SNN was proposed by Bernert \textit{et al.}\cite{Bernert2018Dec} to solve the spike-sorting problem.} Furthermore, Yao \textit{et al.}\cite{yao2021temporal} incorporated a channel attention block into the temporal-wise input of an SNN, as depicted in Fig.~\ref{fig_general}-a, enabling the assessment of frame significance during training and the exclusion of irrelevant frames during inference. Despite employing attention solely in the temporal dimension, this attention mechanism significantly improves the network's performance.

In this paper, we involve both temporal and channel attention mechanisms
in SNNs, which is implemented by efficient 1-D convolution; Fig. \ref{fig_general}-b shows the whole structure, we argue that this cooperative mechanism can enhance the discrimination of the learned features, and can make the temporal-channel learning in SNNs becomes more easier. The main contribution of this work can be summarized as follows:


\begin{enumerate}
    \item We introduce a plug-and-play block into SNNs by considering the temporal and channel attentions cooperatively, which model temporal and channel information in the same phase, achieving better adaptability and bio-interpretability. To the best of our knowledge, this is the first attempt to incorporate the temporal-channel attention mechanism into the most extensively used model, LIF-based SNNs.
    {\item A Cross Convolutional Fusion (CCF) operation with a cross-receptive field is proposed to make use of the associated information. It not only uses the benefit of convolution to minimize parameters but also integrates features from both temporal and channel dimensions in an efficient fashion.
    \item  Experimental results show that TCJA-SNN outperforms previous methods on static and neuromorphic datasets for classification tasks. It also performs well in generation tasks. }

\end{enumerate}

\section{Related Works and Motivation}

\subsection{Training Techniques for Spiking Neural Networks}
\label{sec: bp}

In recent years, the direct application of various ANNs algorithms for training deep SNNs, including gradient-descent-based methods, has gained traction. However, the non-differentiability of spikes poses a significant challenge. The Heaviside function, commonly used to trigger spikes, has a derivative that is zero everywhere except at the origin, rendering gradient-based learning infeasible. To overcome this obstacle, the commonly employed solutions are ANN-to-SNN~\cite{wang2023toward,wu2021tandem,yang2023lc} and the surrogate gradient descent method~\cite{fang2021deep, jin2022sit, Neftci2019Nov, wu2018spatio,rathi2021diet,xie2023event, qiu2023gated, qiu2023vtsnn}.

During the forward pass, the Heaviside function is retained, while a surrogate function replaces it during the backward pass. One simple choice for the surrogate function is the Spike-Operator \cite{Eshraghian2021Sep}, which exhibits a gradient resembling a shifted ReLU function. In our work, we go beyond the conventional surrogate gradient method and introduce two additional surrogate functions: the ATan surrogate function and the triangle-like surrogate function designed by \cite{fang2021incorporating} and \cite{bellec2018long}. These surrogates possess the capability to activate a specific range of samples, making them particularly suitable for the training of deep SNNs. By expanding the repertoire of surrogate functions, we aim to enhance the training process and improve the performance of deep SNNs.


\subsection{Attention Mechanism in Convolutional Neural Networks}

In the realm of ANNs, the Squeeze and Excitation (SE) block, introduced by Hu \textit{et al.} \cite{hu2018squeeze}, has proven to be a highly effective module for enhancing representation. The SE block can be seamlessly incorporated into a network, requiring only a minimal increase in parameters to recalibrate channel information. By employing squeezing and fully connecting operations, it allows the network to learn a trainable scale factor for each channel. This recalibration process significantly improves the discriminative power of individual channels. Recently, Yao \textit{et al.} \cite{yao2021temporal} extended the application of the SE block to SNNs by formulating a temporal-wise attention mechanism. This innovative approach enables SNNs to identify critical temporal frames of interest without being adversely affected by noise or interference. By incorporating temporal-wise attention, the proposed technique achieves state-of-the-art performance across various datasets. This accomplishment serves as compelling evidence for the immense potential of attention mechanisms within SNNs. The utilization of SE blocks and the introduction of temporal-wise attention represent significant advancements in the field of SNNs. These techniques not only enhance the representation capability of SNNs but also offer insights into effectively leveraging attention mechanisms for improved performance. In the Sec. \ref{sec: method}, we aim to explore and further leverage these attention mechanisms to improve the performance of SNNs and unlock their full potential in complex temporal data processing tasks.
\subsection{Motivation}
\begin{figure*}
    \centering
    \includegraphics[width=0.7\textwidth]{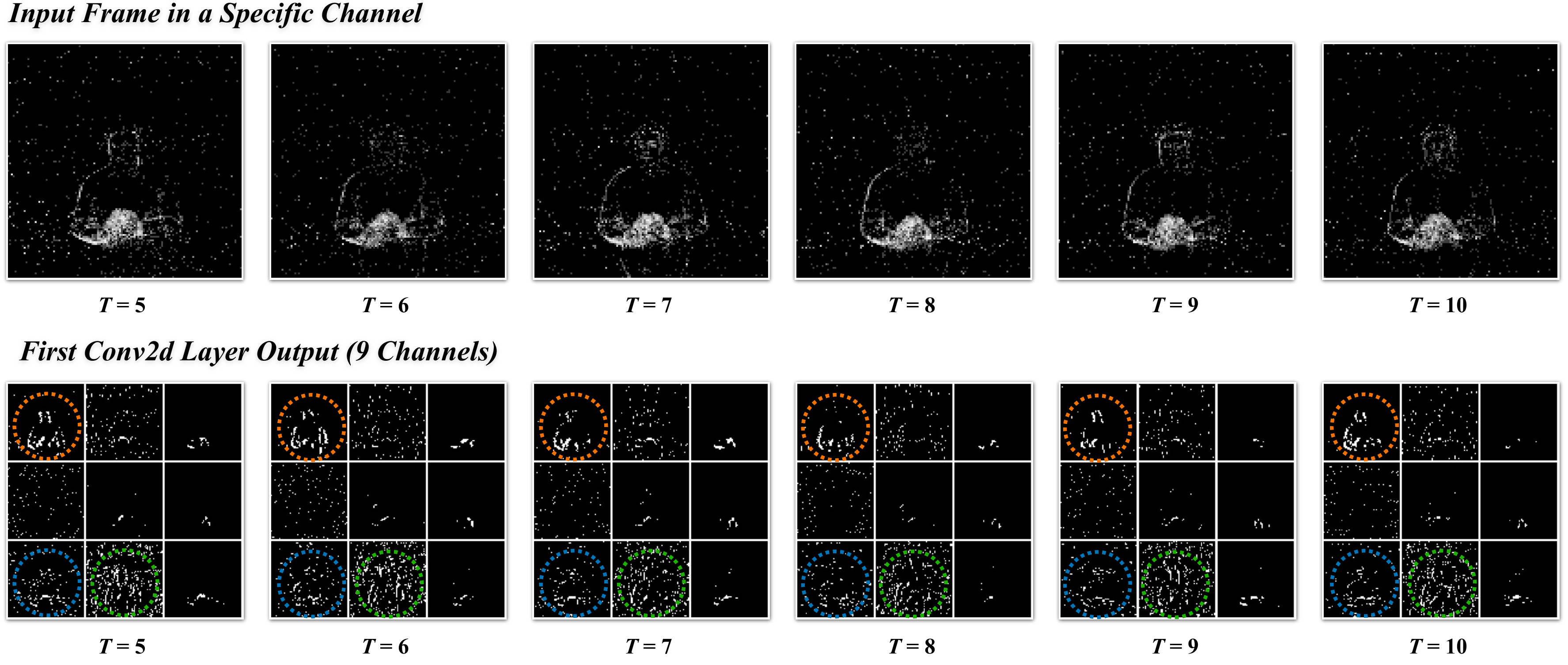}
    \caption{Correlation between proximity time steps and channels. The top row is the input frame selected from DVS128 Gesture dataset. Each figure in the nine-pattern grid of the bottom row denotes a channel output from the first 2-D convolutional layer. It is clear that a significant correlation exists in channels with varying time steps, motivating us to merge the temporal and channel information.}
    \label{fig_motivation}
\end{figure*}

Based on the aforementioned analysis, the utilization of a temporal-wise attention mechanism in SNNs has exhibited substantial progress in effectively processing time-related data streams. Moreover, it has been observed in both biological neural networks~\cite{mante2008functional} and ANNs~\cite{hu2018squeeze} that recalibrating channel features within convolutional layers hold considerable potential for enhancing performance. {Nevertheless, the existing SNNs-based works only process the data with either temporal or channel dimensions, thereby constraining the capacity for joint feature extraction. To illustrate the relationship between temporal steps and channel dimensions, we provide a visual representation. This is achieved by displaying the input frame alongside several adjacent channel outputs, which originate from the initial 2-D convolutional layer, as demonstrated in Fig. \ref{fig_motivation}.} As the circles indicate, a similar firing pattern can be distinguished from the surrounding time steps and channels. To fully use this associated information, we propose the TCJA module, a novel approach for modeling temporal and channel-wise frame correlations. Furthermore, considering the inevitable increases in the model parameters caused by the attention mechanism, we attempt to adopt the 1-D convolution operation to gain a reasonable trade-off between model performance and parameters. 
Furthermore, existing SNN attention mechanisms primarily prioritize classification tasks, neglecting the needs of generation tasks. Our goal is to introduce an attention mechanism that can proficiently handle both classification and generation tasks, thereby establishing a universal attention mechanism for SNNs.



\section{Methodology}
\label{sec: method}
\subsection{Leaky Integrate and Fire Model}
Various spiking neuron models have been proposed to simulate the functioning of biological neurons~\cite{izhikevich2003simple,gerstner2002spiking}, and among them, the Leaky-Integrate-and-Fire (LIF) model~\cite{lapicque1907louis} achieves a commendable balance between simplicity and biological plausibility.
 The membrane potential dynamics of a LIF neuron can be described as~\cite{wu2019direct}:
\begin{equation}
   \tau\frac{dV(t)}{dt} = -(V(t)-V_{reset}) + I(t)
\end{equation}
where $\tau$ denotes a time constant, $V(t)$ represents the membrane potential of the neuron at time $t$, and $I(t)$ represents the input from the presynaptic neurons. For better computational tractability, the LIF model can be described as an explicitly iterative version \cite{roy2019towards}:
\begin{equation}
     \left \{\begin{array}{l}\boldsymbol{V}_{t}^{n}=\boldsymbol{H}_{t-1}^{n} + \frac{1}{\tau}(\boldsymbol{I}_{t-1}^{n} - (\boldsymbol{H}_{t-1}^{n} - \boldsymbol{V}_{reset})) \\
    \boldsymbol{S}_{t}^{n}=\boldsymbol{\Theta}(\boldsymbol{V}_{t}^{n}-\boldsymbol{V}_{threshold})\\
    \boldsymbol{H}_{t}^{n}=\boldsymbol{V}_{t}^{n} \cdot (1-\boldsymbol{S}_{t}^{n}) \\ 

    \end{array} \right.
        \label{eq:SNN layer}
\end{equation}
$\boldsymbol V_t^n$ represents the membrane potential of neurons within the $n$-th layer at time $t$. $\tau$ is a time constant, $\boldsymbol S$ is the spiking tensor with binary value, $\boldsymbol I$ denotes the input from the previous layer,  $\boldsymbol {\Theta(\cdot)}$ denotes the Heaviside step function, $\boldsymbol H$ represents the reset process after spiking. 

As a mainstream neuron model, LIF-based SNN models can be trained directly using surrogate gradient methods~\cite{wu2018spatio} to attain state-of-the-art (SOTA) performance~\cite{deng2021temporal, fang2021incorporating, yao2021temporal}. Moreover, the LIF model is well-suited to common machine-learning frameworks because it allows forward and backward propagation along spatial and temporal dimensions. In our method, the parameters of the LIF model are set as follows:$\tau = 2$, $\boldsymbol V_{reset} = 0$, and $\boldsymbol V_{threshold} = 1$. 

\subsection{Temporal-Channel Joint Attention (TCJA)}


\begin{figure}[htbp]
  \begin{center}
    \includegraphics[width=0.65\linewidth]{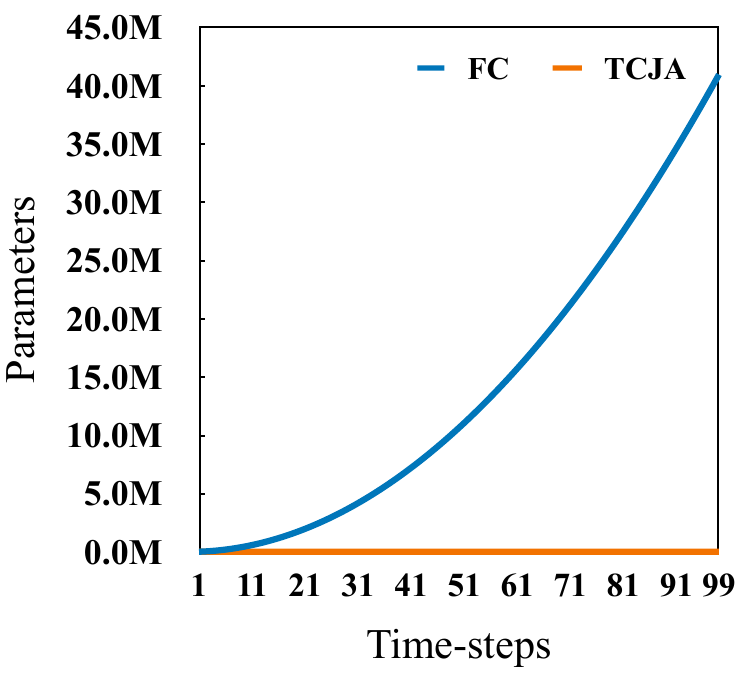}
  \end{center}
  \caption{The growth curve of parameters between Fully-Connected (FC) layer and TCJA layer when channel size $C = 64$.}
  \label{fig_parameters}
\end{figure}
As mentioned above, we contend that the frame at the current time step exhibits a significant correlation with its neighboring frames in both the channel and temporal dimensions. This correlation opens up the possibility of employing a mechanism to establish a connection between these two dimensions. Initially, we employed a fully-connected layer to establish the correlation between the temporal and channel information, as it provides the most direct and prominent connection between these dimensions. However, as the number of channels and time steps increases, the number of parameters grows rapidly with a ratio of $T^2 \times C^2$, as illustrated in Fig.~ \ref{fig_parameters}. our subsequent attempt involved utilizing a 2-D convolutional layer for building this attention mechanism. {Nevertheless, this approach encountered a limitation due to the fixed kernel size, which restricts the receptive field to a confined local area. In conventional CNNs, augmenting the number of layers can expand the receptive field~\cite{simonyan2014very,luo2016understanding}. However, within the context of attention mechanisms, the feasibility of layer stacking, analogous to convolutional networks, is constrained, thereby limiting the receptive field when employing 2D convolutions.} For this reason, it is necessary to decrease the number of parameters while increasing the receptive field. In Sec.~\ref{sec:theoretical}, we provide a detailed theoretical analysis of the receptive field.

\begin{figure*}
    \centering
    \includegraphics[width=0.8\textwidth]{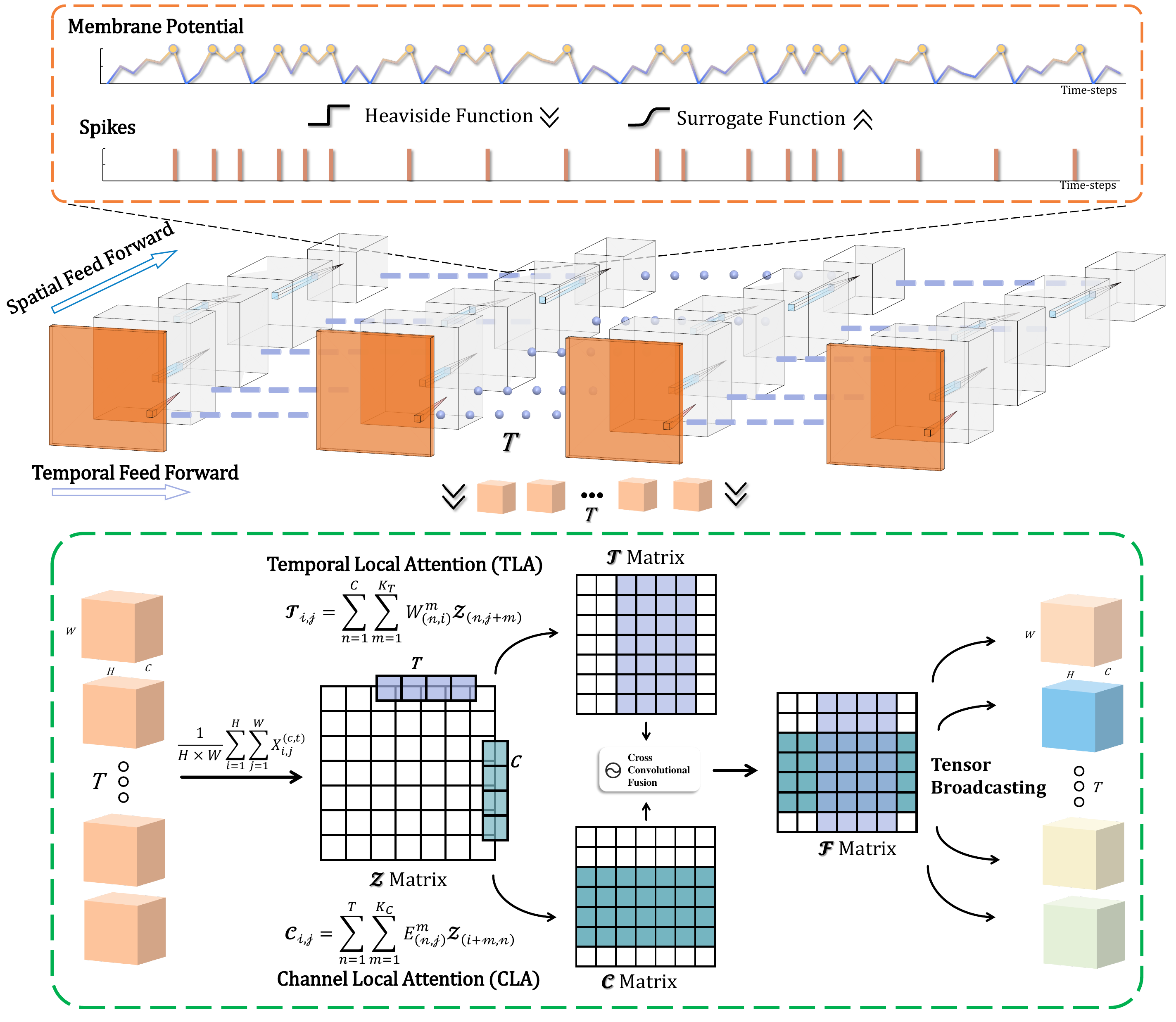}
    \caption{{The Framework of SNN with TCJA module. In SNNs, information is transmitted in the form of spike sequences, encompassing both temporal and spatial dimensions. In temporal-wise, the spiking neuron with a threshold feed-forward in membrane potential ($\boldsymbol{V}$) and spike ($\boldsymbol{S}$) as the Eq. \ref{eq:SNN layer}, and backpropagation with the surrogate function. In spatial-wise, data flows between layers as ANN. The TCJA module operates by initially compressing information along both temporal and spatial dimensions, then apply TLA and CLA to establish the relationship in both temporal and channel dimensions and blend them by CCF layer.}}
    \label{fig_General_TCJA}
\end{figure*}

To effectively incorporate both temporal and channel attention dimensions while minimizing parameter usage and maximizing the receptive field, we present a novel attention mechanism termed Temporal-Channel Joint Attention (TCJA). This attention mechanism is distinguished by its global cross-receptive field and its ability to achieve effective results with relatively fewer parameters, specifically $T^2 + C^2$. Fig. \ref{fig_General_TCJA} shows the overall structure of the proposed TCJA, and we will introduce its key components in detail in the following. In Sec. \ref{sec_squeeze}, we utilize the squeezing operation on the input frame. Next, we introduce the temporal-wise local attention (TLA) mechanism and channel-wise local attention (CLA) mechanism in Sec. \ref{sec_TLA} and Sec. \ref{sec_CLA}, respectively. At last, we introduce the cross-convolutional fusion (CCF) mechanism to conjointly learn the information of temporal and channel in Sec. \ref{sec_CCF}.

\subsubsection{Average Matrix by Squeezing} 
\label{sec_squeeze}
In order to efficiently capture the temporal and channel correlations between frames, we first perform the squeeze step on the spatial feature map of the input frame stream $X\in \mathbb{R}^{T\times H\times W\times C} $, where $C$ denotes the channel size, and $T$ denotes the time step. The squeeze step calculates an average matrix $\mathcal{Z}\in \mathbb{R}^{C\times T}$ and each element $\mathcal{Z}_{(c,t)}$ of the average matrix $\mathcal{Z}$ as:

\begin{equation}
    \label{equ_squeeze}
    \mathcal{Z}_{(c,t)} = \frac{1}{H\times W}\sum_{i=1}^{H}{\sum_{j=1}^{W}{X^{(c,t)}_{i,j}}}
\end{equation}
where $X^{(c,t)}$ is the input frame of $c$-th channel at time step $t$.

\subsubsection{Temporal-wise Local Attention (TLA)}
\label{sec_TLA}

Following the squeeze operation, we propose the TLA mechanism for establishing temporal-wise relationships among frames. We argue that the frame in a specific time step interacts substantially with the frames in its adjacent positions. Therefore, we adopt a 1-D convolution operation to model the local correspondence in the temporal dimension, as shown in Fig. \ref{fig_General_TCJA}. In detail, to capture the correlation of input frames at the temporal level, we perform $C$-channel 1-D convolution on each row of the average matrix $\mathcal{Z}$, and then accumulate the feature maps obtained by convolving different rows of the average matrix $\mathcal{Z}$. The whole TLA process can be described as:
\begin{equation}
    \label{equ_time_attention}
    \mathcal{T}_{i,j} = \sum_{n=1}^{C}{\sum_{m=0}^{K_T-1}{W_{(n,i)}^m \mathcal{Z}_{(n,j+m)}}}
\end{equation}
Here, $K_T$ ($K_T$ \textless $T$) denotes the size of the convolution kernel, which indicates the number of time steps considered for the convolution operation. The parameter $W_{(n,i)}^m$ is a learnable parameter that represents the $m$-th parameter of the $i$-th channel when performing a 1-D convolution operation with $C$ channels on the $n$-th row of the input tensor $\mathcal{Z}$. $\mathcal{T}\in \mathbb{R}^{C\times T}$ is the attention score matrix after the TLA mechanism.

\subsubsection{Channel-Wise Local Attention (CLA)}
\label{sec_CLA}

As aforementioned, the frame-to-frame saliency score should not only take into account the temporal dimension but also take into consideration the information from adjacent frames along the channel dimension. In order to construct the correlation of different frames with their neighbors channel-wise, we propose the CLA mechanism. Similarly, as shown in Fig.~\ref{fig_General_TCJA}, we perform $T$-channel 1-D convolution on each column of the matrix $\mathcal{Z}$, and then sum the convolution results of each row. This process can be described as:
\begin{equation}
    \label{equ_channel_attention}
    \mathcal{C}_{i,j} = \sum_{n=1}^{T}{\sum_{m=0}^{K_C-1}{E_{(n,j)}^m}} \mathcal{Z}_{(i+m,n)}
\end{equation}
where $K_C$ ($K_C$ \textless $C$) represents the size of the convolution kernel, and $E_{(n,i)}^m$ is a learnable parameter, representing the $m$-th parameter of the $i$-th channel when performing $T$-channel 1-D convolution on $n$-th column of $\mathcal{Z}$. $\mathcal{C}\in \mathbb{R}^{C\times T}$ is the attention score matrix after CLA mechanism. 

{To maintain dimensional consistency between the input and output, a ``same padding" technique is employed in both the TLA and CLA mechanisms. This padding strategy ensures that the output dimension matches the input dimension by adding an appropriate number of zeros to the input data. Specifically, this technique involves padding the input array with zeros on both sides, where the number of zeros added is determined based on the kernel size and the stride.}

\subsubsection{Cross Convolutional Fusion (CCF)}
\label{sec_CCF}
After TLA and CLA operations, we get the temporal (TLA matrix $\mathcal{T}$) and channel (CLA matrix $\mathcal{C}$) saliency scores of the input frame and its adjacent frames, respectively. Next, to learn the correlation between temporal and channel frames in tandem, we propose a cross-domain information fusion mechanism, \textit{i.e.}, the CCF layer. The goal of CCF is to calculate a fusion information matrix $\mathcal{F}$, and $\mathcal{F(\textit{i}, \textit{j})}$ is used to measure the potential correlation between the $i$-th channel of the $j$-th input temporal frame and other frames. Specifically, the joint relationship between frames can be obtained by performing an element-wise multiplication of $\mathcal{T}$ and $\mathcal{C}$ as follows:
\begin{equation}
    \label{equ_ccf}
    \begin{aligned}
        \mathcal{F}_{i,j} =  \boldsymbol{\sigma}( \mathcal{T}_{i,j} \cdot \mathcal{C}_{i,j}) =
        \boldsymbol{\sigma} ( \sum_{n=1}^{C}{\sum_{m=0}^{K_T-1}{W_{(n,i)}^m \mathcal{Z}_{(n,j+m)}}}\cdot \\ \sum_{n=1}^{T}{\sum_{m=0}^{K_C-1}{E_{(n,j)}^m \mathcal{Z}_{(i+m,n)}}})
    \end{aligned}
\end{equation}
\begin{figure*}[htbp]
    \centering
    \includegraphics[width=0.8\textwidth]{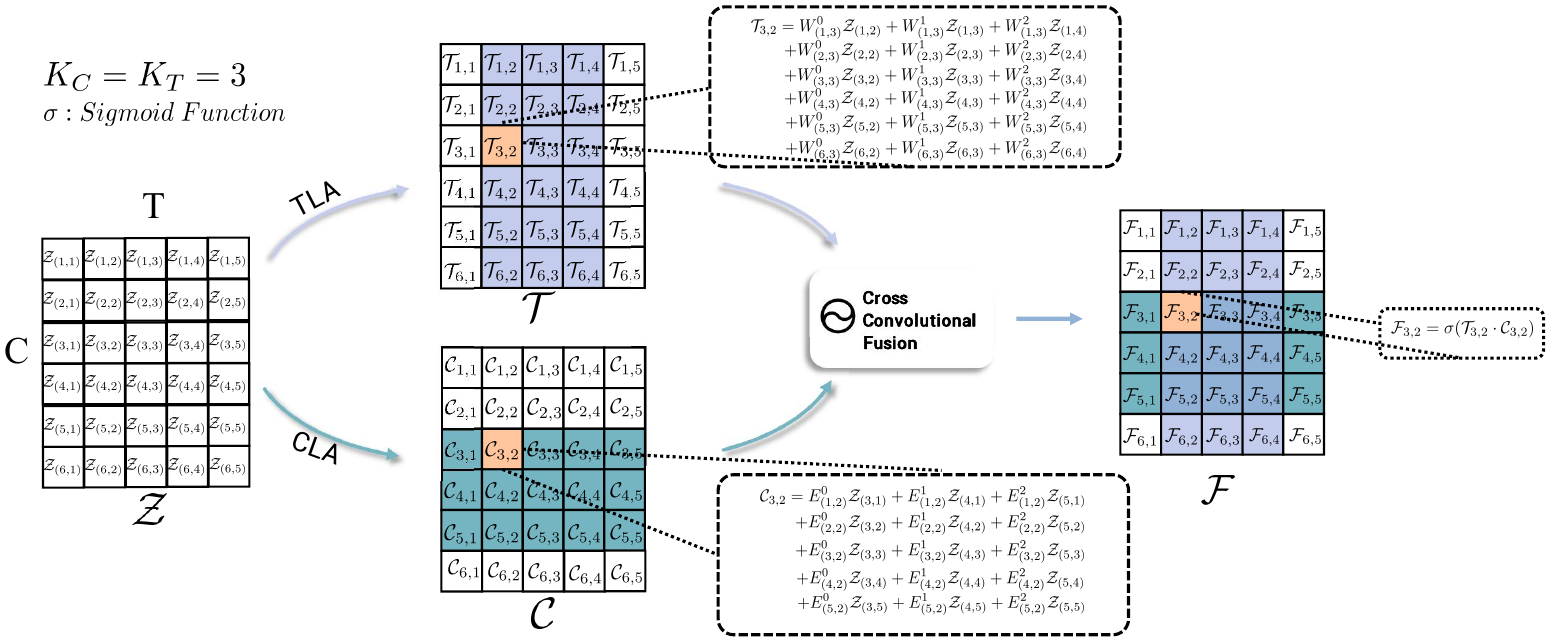}
    \caption{Illustration of the proposed TCJA. We give an average matrix $\mathcal{Z} \in \mathbb{R}^{6\times 5}$, and the goal of TCJA is to calculate a fusion matrix $\mathcal{F}$ integrating temporal and channel information. For instance, for a specific element in $\mathcal{F}$: $\mathcal{F}_{3,2}$, its calculation pipeline is as follows: 1) Calculate $\mathcal{T}_{3,2}$ through TLA mechanism (Eq. \ref{equ_time_attention}); 2) Utilize CLA mechanism (Eq. \ref{equ_channel_attention}) to calculate $\mathcal{C}_{3,2}$, and the calculation results are shown in the black dotted box in the figure; 3) Adopt CCF mechanism (Eq. \ref{equ_ccf}) to jointly learn temporal and channel information to obtain $\mathcal{F}_{3,2}$. In addition, we can also find that after the CCF mechanism, $\mathcal{F}_{3,2}$ integrates the information of the elements in the cross receptive field (Colored areas in $\mathcal{F}$) as the anchor point, which indicates the \emph{Cross} Convolutional Fusion.}
    \label{fig_TCJA_demo}
\end{figure*}
where $\boldsymbol{\sigma}$ represents the Sigmoid function. Fig.~\ref{fig_TCJA_demo} is provided to enhance the understanding of the entire computational process. 

\subsection{Training Framework}
We integrate the TCJA module into the existing benchmark SNNs and propose the TCJA-SNN. Since the process of neuron firing is non-differentiable, we utilize the derived ATan surrogate function $\sigma'(x) = \frac{\alpha}{2(1 + (\frac{\pi}{2}\alpha x)^2)}$ and the derived triangle-like surrogate function $\epsilon'(x) = \frac{1}{\gamma^2}\text{max}(0,\gamma-|{x}-1|)$ for backpropagation, which is proposed by \cite{fang2021incorporating} and \cite{bellec2018long}, respectively. {This latter function is particularly applied in the TCJA-TET-SNN, in alignment with the default surrogate function specification for TET-based architectures.} In our method, the Spike Mean-Square-Error (SMSE) \cite{fang2021deep, fang2021incorporating} is chosen as the loss function, which can be expressed as:
\begin{equation}
    \mathcal L =\frac{1}{T} \sum^{T-1}_{t = 0} \mathcal L_{t} = \frac{1}{T} \sum^{T-1}_{t = 0} \frac{1}{E} \sum^{E-1}_{i = 0} (s_{t, i} - g_{t,i})^2
\end{equation}
where $T$ denotes the simulation time step, $E$ is the number of labels, $s$ represents the network output and $g$ represents the one-hot encoded target label.
We also employ the Temporal Efficient Training (TET)~\cite{deng2021temporal} loss, which can be represented as:
\begin{equation}
    \mathcal{L} = \frac{1}{T}\cdot\sum_{t=1}^{T}\mathcal{L}_\text{CE}[s(t),g(t)]
\end{equation}
where $T$ is the total simulation time, $\mathcal{L}_\text{CE}$ denotes the cross-entropy loss, $s$ is the network output, and $g$ represents the target label. The cross entropy loss here can be represented by:
\begin{equation}
    \mathcal{L}_\text{CE}(p, y) = -\sum_{c=1}^{M} y_{o,c} \log(p_{o,c})
\end{equation}
where $M$ is the number of classes, $y_{o,c}$ is a binary indicator (0 or 1) if class label $c$ is the correct classification for observation $o$, $p_{o,c}$ is the predicted probability of observation $o$ being of class $c$.

To estimate the classification accuracy, we define the predicted label $l_{p}$ as the index of the neuron with the highest firing rate $l_{p} = \mathop{\max_{i}} \frac{1}{T}\sum_{t=0}^{T-1} s_{t, i}$. Since the TCJA module simply utilizes the 1-D convolutional layer and Sigmoid function, it can be effortlessly introduced into the current network architecture as a plug-and-play module without adjusting to backpropagation.

\section{Experiments}
We evaluate the classification performance of TCJA-SNN on both neuromorphic datasets (CIFAR10-DVS, N-Caltech 101, and DVS128 Gesture) and static datasets (Fashion-MNIST, CIFAR10, CIFAR100). Note that all neuromorphic datasets are collected from the event sensor. To verify the effectiveness of the proposed method, we integrate the TCJA module into several architectures \cite{deng2021temporal,fang2021incorporating} with competitive performance to see if the integrated architecture can generate significant improvement. 

\subsection{Dataset}
\subsubsection{Dataset Introduction}

We have conducted experiments on both event-stream and static datasets for object classification. The summaries of the datasets involved in the experiment are listed below.

\begin{itemize}
    \item \textbf{CIFAR10-DVS} The CIFAR10-DVS \cite{li2017cifar10dvs} dataset is an adapted event-driven version from the popular static dataset CIFAR10. This dataset converts 10,000 frame-based images of 10 classes into event streams with the dynamic vision sensor. Since the CIFAR10-DVS dataset does not divide training and testing sets, we split the dataset into 9k training images and 1k test images and reduced the spatial resolution from $128\times128$ to $48\times48$ as \cite{deng2021temporal,li2021differentiable,li2022neuromorphic,meng2022training}.
    \item \textbf{N-Caltech 101} The N-Caltech 101 \cite{orchard2015nmnist} dataset is also converted from the original version of Caltech 101 \cite{1384978} with a slight change in object classes to avoid confusion. The N-Caltech 101 consists of 100 object classes plus one background class. We apply the 9: 1 train-test split as CIFAR10-DVS.
    \item \textbf{DVS128 Gesture} The DVS128 Gesture \cite{amir2017dvsg} dataset is an event-stream dataset composed of 11 kinds of hand gestures from 29 subjects under three different illumination conditions, directly captured with the DVS128 camera. In this paper, we employs all 11 gesture categories for the purpose of classification.
    \item \textbf{Fashion-MNIST} The Fashion-MNIST \cite{xiao2017fashionmnist} is a tiny but demanding static dataset designed to serve as a straight replacement for the original MNIST dataset for more complicated visual patterns. The Fashion-MNIST dataset contains 70,000 grayscale images of 10 kinds of fashion products, all in a $28\times 28$ size.
    {\item \textbf{CIFAR10/100} The CIFAR10/100 dataset~\cite{krizhevsky2009cifar10} consists of 60,000 32 $\times$  32 images with 3 channels in 10/100 classes. There are 50,000 training images and 10,000 testing images. }
\end{itemize}

\subsubsection{Neuromorphic Dataset Preprocessing}
We use the integrating approach to convert event stream to frame data, which is commonly used in SNNs\cite{fang2021incorporating,kugele2020efficient,wu2019direct,yao2021temporal}, to preprocess neuromorphic datasets. The coordinate of an event can be described as:
\begin{equation}
    E(x_{i}, y_{i}, p_{i})
\end{equation}
where $x_{i}$ and $y_{i}$ event’s coordinate, $p_{i}$ represents the event. In order to reduce computational consumption, we group events into $T$ slices, where $T$ is the network's time simulation step. {A frame in the integrated frame data, denoted as $F(j)$, refers to the pixel value at position $(p, x, y)$, represented as $F(j, p, x, y)$. It is obtained by integrating events indexed between $j_{l}$ and $j_{r}$ from the event stream, where  Here, $j_l$ represents the initial timestamp for accumulation, and $j_r$ denotes the final timestamp.} The process can be described as:
\begin{equation}
    \begin{split}j_{l} & = \left\lfloor \frac{N}{T}\right \rfloor \cdot j \\
    j_{r} & = \begin{cases} \left \lfloor \frac{N}{T} \right \rfloor \cdot (j + 1) & \text{if}~~ j <  T - 1 \cr N &  \text{if} ~~j = T - 1 \end{cases} \\
F(j, p, x, y) &= \sum_{i = j_{l}}^{j_{r} - 1} \mathcal{I}_{p, x, y}(p_{i}, x_{i}, y_{i})\end{split}
\end{equation}
where $\lfloor \cdot \rfloor$ is the floor operation, and $\mathcal{I}_{p, x, y}(p_{i}, x_{i}, y_{i})$ is an indicator function and it equals 1 only when $(p, x, y) = (p_{i}, x_{i}, y_{i})$. {The function $F(j)$ is primarily designed to accumulate event data within a specified range. This accumulation is then segmented into frames, facilitating a format that is more conducive to the simulation of SNNs. This structured approach in framing the data not only enhances the compatibility with SNNs but also enables a more efficient analysis and processing of the event data, aligning it with the inherent temporal dynamics of SNNs.}
\subsubsection{Data Augmentation}
To mitigate the apparent overfitting on the CIFAR10-DVS dataset, we adopt the neuromorphic data augmentation, which is also used in \cite{deng2021temporal,li2021differentiable,li2022neuromorphic,meng2022training} for training the same dataset. We follow the same augmentation setting as \cite{li2022neuromorphic}: we utilize horizontal Flipping and Mixup \cite{zhang2018mixup} in each frame, where the probability of Flipping is set to 0.5, and the Mixup interpolation factor is sampled from a beta distribution where $\alpha = 0.5, \beta = 0.5$. Then, we randomly select one augmentation among Rolling, Rotation, Cutout, and Shear, where random Rolling range is 5 pixels, the degree of Rotation is sampled from the uniform distribution where $\alpha=-15, \beta=15$, the side length of Cutout is sampled from the uniform distribution where $\alpha=1, \beta=8$, and the shear degree is also sampled from the uniform distribution where $\alpha=-8, \beta=8$.
\subsection{Network Architecture}
{The architectures of networks corresponding to various datasets are enumerated in Table \ref{tab:arch}. In the construction of each network, Kaiming~\cite{he2015delving} initialization is methodically applied to both convolutional and fully-connected layers.}
\begin{table*}[h]
\centering
\caption{{The network architecture setting for each dataset. $x$C$y$/MP$y$/AP$y$ denotes is the Conv2D/MaxPooling/AvgPooling layer with output channels = $x$ and kernel size = $y$. $n$FC denotes the fully connected layer with output feature = $n$, $m$DP is the spiking dropout layer with dropout ratio $m$. The Voting layer is a 1-D average pooling layer.}}
\begin{tabular}{ll}

\toprule
\textbf{Dataset}        & \textbf{Network Architecture}                                     \\                                                  \midrule
DVS128 Gesture & \begin{tabular}[c]{@{}l@{}}128C3-LIF-MP2-128C3-LIF-MP2-128C3-LIF-MP2-128C3-LIF\\ -MP2-128C3-LIF-MP2-0.5DP-512FC-LIF-0.5DP-100FC-LIF-Voting\end{tabular}  \\ \midrule
CIFAR10-DVS    & \begin{tabular}[c]{@{}l@{}}64C3-LIF-128C3-LIF-AP2-256C3-LIF-256C3-LIF-AP2-512C3-LIF\\ -512C3-LIF-AP2-512C3-LIF-512C3-LIF-AP2-10FC-LIF\end{tabular} \\ \midrule
N-Caltech 101  & \begin{tabular}[c]{@{}l@{}}64C3-LIF-MP2-128C3-LIF-MP2-256C3-LIF-MP2-256C3-LIF\\ -MP2-512C3-LIF-0.8DP-1024FC-LIF-0.5DP-101FC-LIF\end{tabular}             \\ \midrule

Fashion-MNIST  & 128C3-LIF-AP2-128C3-LIF-AP2-0.5DP-512FC-LIF-0.5DP-10FC-LIF                                                                                               \\ \midrule
{CIFAR10/100}   & \begin{tabular}[c]{@{}l@{}}64C7-{LIF-64C3-LIF-64C3}*2-{LIF-128C3-LIF-128C3}*2\\-{LIF-256C3-LIF-256C3}*2-{LIF-512C3-LIF-512C3}*2-LIF-10/100FC\end{tabular}                                                                                              \\ \bottomrule
\end{tabular}
\label{tab:arch}
\end{table*}
\begin{table*}[htbp]
    \centering
    \caption{Hyperparameter settings of TCJA-SNN.}
    \begin{tabular}{lcccccc}
        \toprule 
         \textbf{Hyperparameter} & \textbf{CIFAR10-DVS} & \textbf{N-Caltech 101} & \textbf{DVS128} & \textbf{Fashion-MNIST} & {\textbf{CIFAR10}} & {\textbf{CIFAR100}} \\
         \midrule
         Optimizer & Adam & Adam & Adam & Adam& {SGD}& {SGD}\\
         Learning Rate & $1e-3$  & $1e-3$ & $1e-3$ & $1e-3$ & {$1e-1$}& {$1e-1$} \\
         Batch Size & 64 & 32 & 16 & 128& {128}& {128}\\
         $T$ & 10 & 14 & 20 & 8& {6/4}& {6/4}\\
         Automatic Mixed Precision & \xmark & \cmark & \cmark & \xmark& {\xmark}& {\xmark}\\
         Training Epochs & 1000 & 500 & 1000 & 1000& {250} & {250}\\
         \bottomrule
    \end{tabular}
    \label{tab:hyperparameters}
\end{table*}
For the DVS128 dataset, we utilize the same network structure and hyper-parameters as the \cite{fang2021incorporating} and add the TCJA module before the last two pooling layers. Dropout (DP) \cite{srivastava2014dropout} rate is set to 0.5 in accordance with the original network. {We add a 1-D average pooling voting layer in the last layer, which yielded a 10-dimensional vector as the vote outcome. This is because the preprocess of DVS128 Gesture simulates a longer timestep ($T$=20), through such a voting layer the robustness of the network can be improved \cite{wu2019direct}. }

For the CIFAR10-DVS dataset, we adopt the VGG11-like architecture introduced in TET \cite{deng2021temporal}.  Due to the significant overfitting, we adopt the data augmentation as \cite{li2022neuromorphic} and \cite{deng2021temporal}. To maintain the same training settings as \cite{deng2021temporal} for TCJA-TET-SNN, we use the triangle surrogate function, eliminate the last LIF layer, and replace the SMSE loss with TET loss. For TCJA-SNN, the TCJA module is added before the last two pooling layers, and for TCJA-TET-SNN, the TCJA module is included before the first pooling layer as the replacement of surrogate function and loss.

For N-Caltech 101 dataset, we combine two architectures together and add the TCJA module before the last two pooling layers. We first reserve a pooling for each layer; then, with the network going deeper, spatial resolution is reduced with the increasing channel number. To relieve the evident overfitting, the ratio of the first dropout layer is increased to 0.8.

For the Fashion-MNIST dataset, we follow the network structure from \cite{fang2021incorporating}. Note that the first convolutional layer is a static encoding layer, transforming the static image into spikes.

{For the CIFAR10/100 dataset, we employ the MS-ResNet architecture, as detailed in Ref.~\cite{hu2021advancing}, to validate the effectiveness of the TCJA on deep residual neural networks. Specifically, we utilize the standard MS-ResNet-18 architecture for classifying the CIFAR datasets. The TCJA module is integrated at the bottom of each MS-ResNet block.}
\subsection{Network Implementation}
\label{sec:impl}
We train and test our method on a workstation equipped with two Tesla P4 and two Tesla P10 GPUs. As the memory consumption, we use the Tesla P10 to train and test the CIFAR10-DVS dataset, N-Caltech 101 dataset, and DVS128 Gesture dataset, and use the Tesla P4 to train and test the Fashion-MNIST, CIFAR10/100 dataset. In the various datasets under consideration, the hyperparameters are detailed in Table \ref{tab:hyperparameters}. The learning rate has been empirically set to $1 \times 10^{-3}$ for each dataset when utilizing the Adam optimizer. Conversely, for the implementation involving ResNet with the SGD optimizer, a higher learning rate of $1 \times 10^{-1}$ has been employed, as the SGD optimizer necessitates a more substantial learning rate. and train the same epochs as \cite{fang2021incorporating} except N-Caltech 101, which is not tested in \cite{fang2021incorporating}. we enable the automatic mixed precision in N-Caltech 101 and DVS128 Gesture for the excessive resolution ($180 \times 240$ and $128 \times 128$). {We strategically detach the reset process during backpropagation, a technique increasingly recognized for its effectiveness in optimizing SNNs. The detach operation decouples the reset operation from the computational graph. Such a detachment has been empirically validated to enhance network performance, offering a more efficient approach to managing the dynamics of SNNs~\cite{fang2021deep, fang2021incorporating, stromatias2015robustness}. This process ensures that the essential learning dynamics are retained while unnecessary computational complexities are minimized, thereby improving the overall efficacy of the network.}
\subsection{Comparison with Existing Classification SOTA Works}
The performance of two TCJA-SNN variants is compared with some SOTA models in Tab. \ref{tab:main}, Tab.~\ref{tab:CIFAR} and Tab. \ref{tab_fashion}. We train and test two variants with SpikingJelly \cite{SpikingJelly} package based on PyTorch \cite{paszke2019pytorch} framework, resulting in enhanced performance across all tasks.
Some studies \cite{fang2021deep,wu2021liaf,yao2021temporal} substitute binary spikes with floating-point spikes in whole or in part and retain the same temporal forward pipeline as SNN to obtain improved classification accuracy. 
Thus, we devise two variants to validate the efficiency of TCJA-SNN by utilizing the TET loss function. On CIFAR10-DVS, we obtain 1.7\% advantage over the prior method with binary spikes.
 on the N-Caltech 101 dataset, we achieved a classification accuracy of 82.5\%, surpassing previous work by 1.6\%.  On DVS128, we get an accuracy of 99.0\%, which is higher than TA-SNN \cite{yao2021temporal} using three times fewer simulation time steps. Furthermore, by using a basic 7-layer CNN on the static dataset Fashion MNIST, our method can achieve the highest classification accuracy with the fewest simulation time steps. {In the context of the CIFAR10 and CIFAR100 datasets, the implementation of TCJA demonstrates a significant improvement over the baseline models~\cite{hu2021advancing} that do not incorporate TCJA. Specifically, there is an enhancement of 2.08\% and 3.83\% in classification accuracy for CIFAR10 and CIFAR100, respectively. Additionally, our method surpasses the current SOTA models, as referenced in Ref.~\cite{yaoglif}, by margins of 0.84\% for CIFAR10 and 0.93\% for CIFAR100.}
Overall, with binary spikes, TCJA-SNN simulates no-more time steps while getting higher performance. Furthermore, our method can achieve higher classification accuracy by adopting the non-binary spike technique.

\begin{table*}[t]
   
   \centering
   \caption{The comparison between the proposed methods and existing SOTA techniques on three mainstream neuromorphic datasets. (Bold: the best)}
   \begin{adjustbox}{max width=\linewidth}
   \begin{tabular}{lccccccc}
   
   \toprule 
   \multirow{3}*{{Method}} &
   \multirow{3}*{{Binary Spikes}} & \multicolumn{2}{c}{{CIFAR10-DVS}} & \multicolumn{2}{c}{{N-Caltech 101}} & \multicolumn{2}{c}{{DVS128}} \\
   \cmidrule(l{2pt}r{2pt}){3-4}\cmidrule(l{2pt}r{2pt}){5-6}\cmidrule(l{2pt}r{2pt}){7-8}
    &{} & {$T$ Step} & {Acc.} & {$T$ Step} & {Acc.} & {$T$ Step} & {Acc.} \\
   \midrule
   SLAYER~\cite{shrestha2018slayer}\textsuperscript{NeurIPS-2018} & \cmark & - & - & - & - & 1600 & 93.4 \\
   HATS~\cite{sironi2018hats}\textsuperscript{CVPR-2018} & N/A & N/A & 52.4 & N/A & 64.2 & - & - \\ 
   DART~\cite{ramesh2019dart}\textsuperscript{TPAMI-2019} & N/A & N/A & 65.8 & N/A & 66.8 & - & - \\ 
   NeuNorm~\cite{wu2019direct}\textsuperscript{AAAI-2019} & \cmark & 230-292 & 60.5 & - & - & - & - \\ 
   Rollout~\cite{kugele2020efficient}\textsuperscript{Front. Neurosci-2020} & \cmark & 48 & 66.8 & - & - & 240 & 97.2 \\ 
   DECOLLE~\cite{kaiser2020synaptic}\textsuperscript{Front. Neurosci-2020} & \cmark & - & - & - & - & 500 & 95.5 \\ 
   LIAF-Net~\cite{wu2021liaf}\textsuperscript{TNNLS-2021} & \xmark& 10 & 70.4 & - & - & 60 & 97.6 \\ 
   tdBN~\cite{zheng2021going}\textsuperscript{AAAI-2021} & \cmark& 10 & 67.8 & - & - & 40 & 96.9 \\ 
   PLIF~\cite{fang2021incorporating}\textsuperscript{ICCV-2021} & \cmark& 20 & 74.8 & - & - & 20 & 97.6 \\ 
   TA-SNN~\cite{yao2021temporal}\textsuperscript{ICCV-2021} & \xmark& 10 & 72.0 & - & - & 60 & 98.6 \\ 
   SEW-ResNet~\cite{fang2021deep}\textsuperscript{NeurIPS-2021} & \cmark& 16 & 74.4 & - & - & 16 & 97.9 \\ 
   Dspike~\cite{li2021differentiable}\textsuperscript{NeurIPS-2021} & \cmark& 10 & 75.4$^*$ & - & - & - & - \\ 
   SALT~\cite{kim2021optimizing}\textsuperscript{Neural Netw-2021} & \cmark & 20 & 67.1 & 20 & 55.0 & - & - \\ 
   TET~\cite{deng2021temporal}\textsuperscript{ICLR-2022} &\xmark & 10 & 83.2$^*$ & - & - & - & - \\ 
   DSR~\cite{meng2022training}\textsuperscript{CVPR-2022} &\cmark & 10 & 77.3$^*$ & - & - & - & - \\ {Event Transformer}~\cite{li2022event} &\xmark & N/A & 71.2 & N/A & 78.9 & - & - \\ 
   {STCA-SNN}~\cite{wu2023stca}\textsuperscript{Front. Neurosci-2023} &\cmark & 10 & 81.6$^*$ & 14 & 80.9 & - & - \\ 
   
   \midrule
   \textbf{TCJA-SNN} & \cmark & 10 & 80.7$^*$ & 14 & 78.5 & 20 & \textbf{99.0} \\ 
   \textbf{TCJA-TET-SNN} & \xmark & 10 & \textbf{83.3}$^*$ & 14 & \textbf{82.5} & 20 & 98.2 \\
   \bottomrule
   \multicolumn{8}{l}{$^*$ With Data Augmentation.}
   \end{tabular}
   \end{adjustbox}

\label{tab:main}
\end{table*}
\begin{table*}[!t]
   {
   \centering
   \caption{The comparison between the proposed methods and existing SOTA techniques on static CIFAR datasets. (Bold: the best)}
   \begin{adjustbox}{max width=\linewidth} 
   \begin{tabular}{lccccc}
   \toprule 
   \multirow{1}*{{Methods}} &
     \multirow{1}*{{Architecture}} & \multicolumn{2}{c}{{CIFAR10}} & \multicolumn{2}{c}{{CIFAR100}} \\
   \cmidrule(l{2pt}r{2pt}){3-4}\cmidrule(l{2pt}r{2pt}){5-6}
     & {} & {$T$ Step} & {Acc.} & {$T$ Step} & {Acc.} \\
    \midrule
ANN2SNN \cite{hao2023reducing}\textsuperscript{AAAI-2023}                 & ResNet-18/ResNet-20                    & 32               & 95.42 & 32 &  65.50               \\
tdBN  \cite{zheng2021going}\textsuperscript{AAAI-2021}                 & Spiking-ResNet-19                     & 6               & 93.16 & 6& 71.12                \\

TET \cite{deng2021temporal}\textsuperscript{ICLR-2022}                & Spiking-ResNet-19                    & 6               & 94.50 &6 & 74.72                \\
    
RecDis \cite{guo2022recdis}\textsuperscript{CVPR-2022}                  & Spiking-ResNet-19                    & 6               & 94.71 &6 & 74.10                \\
   
GLIF \cite{yaoglif}\textsuperscript{NeurIPS-2022}                &Spiking-ResNet-19                     & 6              & 95.03 & 6&  77.35               \\
\midrule
 \text{MS-ResNet} \cite{hu2021advancing}              & MS-ResNet-18                     & 6               & 93.79 & 6&  74.45               \\
  
    \cmidrule{2-6}
\multirow{2}{*}{\textbf{TCJA-SNN}}       & MS-ResNet-18      
& 6 &\textbf{95.87}&6& \textbf{78.28}    \\
         &   MS-ResNet-18    & 4                   &95.60 &4      &77.72 \\                       
     \midrule
      \text{ANN} \cite{hu2021advancing}              & MS-ResNet-18                     & N/A              & 96.41 & N/A& 80.67                \\
    \bottomrule
      \end{tabular}
\end{adjustbox}

\label{tab:CIFAR}
}
\end{table*}

\begin{table}[t]
    \centering
\caption{Static Fashion-MNIST accuracy.}
    \begin{adjustbox}{max width=\linewidth}
   \begin{tabular}{lccc}
   \toprule 
   {{Method}} & {Binary Spike} & {Time Step} & { Accuracy} \\
   \midrule
   ST-RSBP~\cite{zhang2019spike}\textsuperscript{NeurIPS-2019} & \cmark & 400 & 90.1 \\
   LISNN~\cite{cheng2020lisnn}\textsuperscript{IJCAI-2020} & \cmark & 20 & 92.1 \\
   PLIF~\cite{fang2021incorporating}\textsuperscript{ICCV-2021} & \cmark& 8 & 94.4 \\
   \midrule
   \textbf{TCJA-SNN} & \cmark & 8 & \textbf{94.8} \\
   \textbf{TCJA-TET-SNN} & \xmark & 8 & 94.6\\
   \bottomrule
   \end{tabular}
\label{tab_fashion}
\end{adjustbox}
\end{table}
\vspace{15pt}

\subsection{Comparison with Existing Image Generation Works }

\begin{figure}
    \centering
    \includegraphics[width=0.9\linewidth]{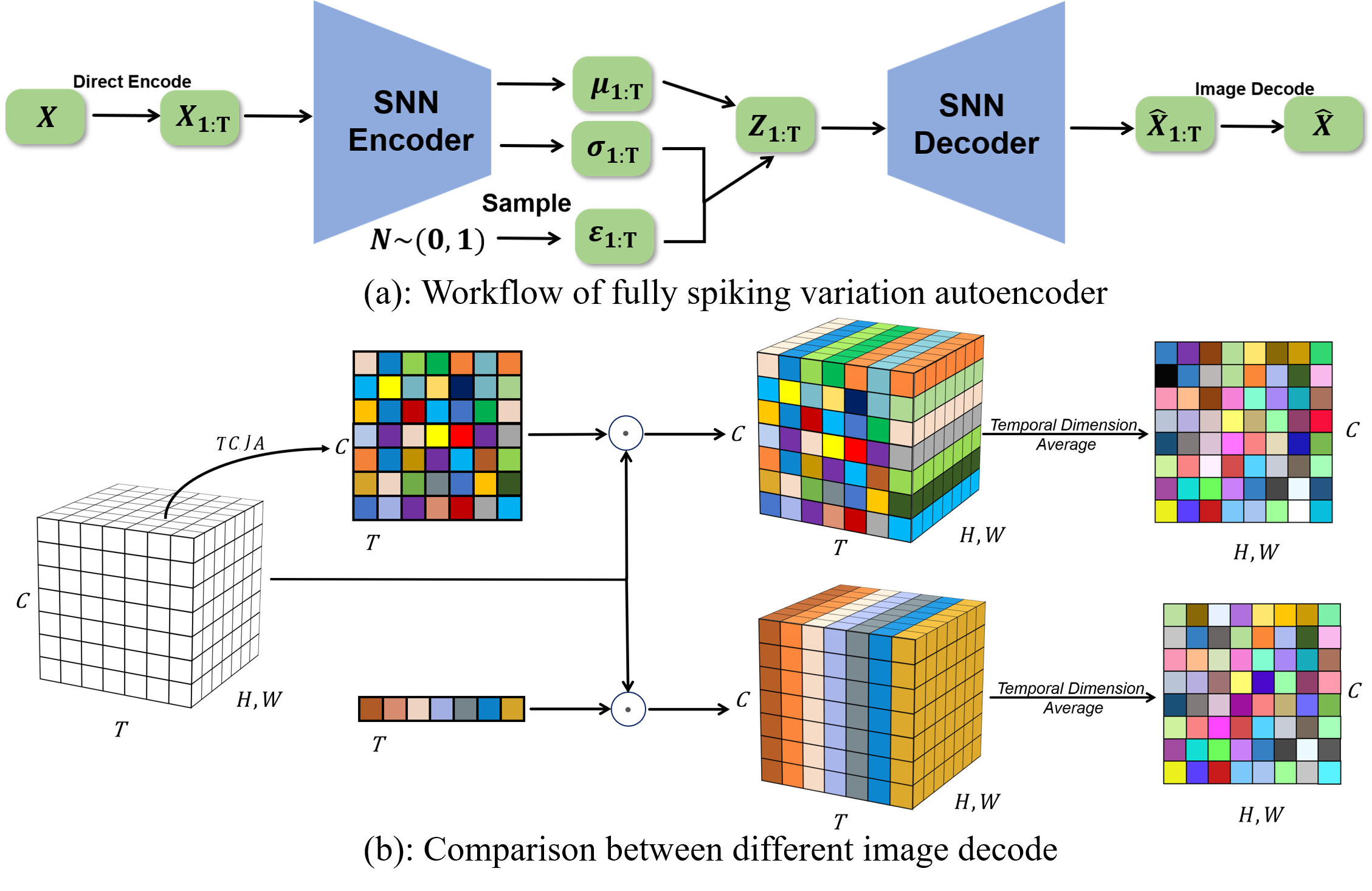}
    \caption{ Architecture of FSVAE and how TCJA applied on it. During training, input images are encoded into spiking inputs, obtaining features $\mu_{1:T}$, $\sigma_{1:T}$ after an SNN encoder. Latent encoding $z_{1:T}$ is randomly generated with a normal distribution. Finally, output images can be reconstructed through a symmetric SNN decoder. By making the best of the abundant temporal information of output spikes, our TCJA image decode performs better.}
    \label{qxr1}
\end{figure}

In this experiment, we build a fully spiking variation autoencoder (FSVAE) for image generation with TCJA. Moreover, we replace the original image decoding way with our novel method by calculating the average output on the temporal dimension after the TCJA block.  The workflow chart of this FSVAE with TCJA applied on image decoding is shown in Fig. \ref{qxr1}. And, we apply log-likelihood evidence lower bound (ELBO) as the loss function:
\begin{equation}
    \label{eq1}
    \begin{aligned}
E L B O= & \mathbb{E}_{q\left(\boldsymbol{z}_{1: T} \mid \boldsymbol{x}_{1: T}\right)}\left[\log p\left(\boldsymbol{x}_{1: T} \mid \boldsymbol{z}_{1: T}\right)\right] \\
& -\operatorname{KL}\left[q\left(\boldsymbol{z}_{1: T} \mid \boldsymbol{x}_{1: T}\right) \| p\left(\boldsymbol{z}_{1: T}\right)\right],
\end{aligned}
\end{equation}
where the first term is the reconstruction loss between the original input and the reconstructed one, which is the mean square error (MSE) in this model. The second term is the Kullback-Leibler (KL) divergence, representing the closeness of prior and posterior.
\\

\begin{figure}
    \centering
    \includegraphics[width=0.9\linewidth]{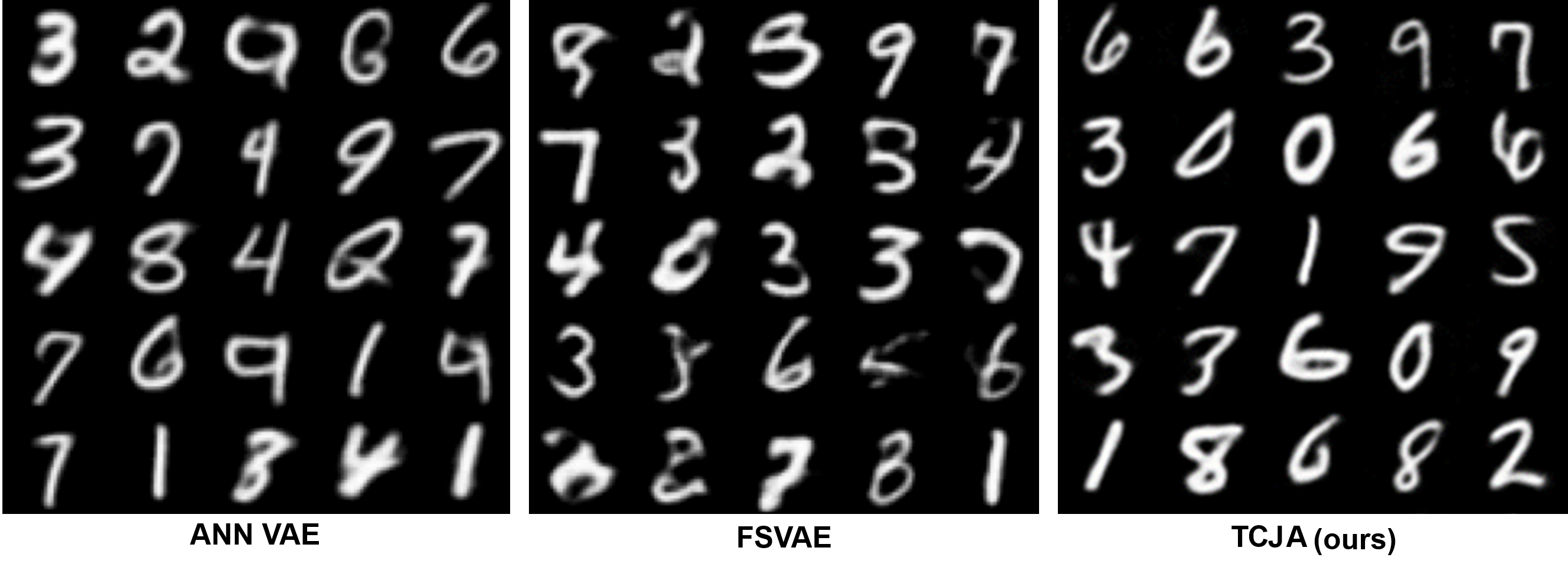}
    \caption{Generated images of ANN VAE, FSVAE, and ours TCJA on MNIST dataset.}
    \label{qxr2}
\end{figure}

\begin{figure}
    \centering
    \includegraphics[width=0.9\linewidth]{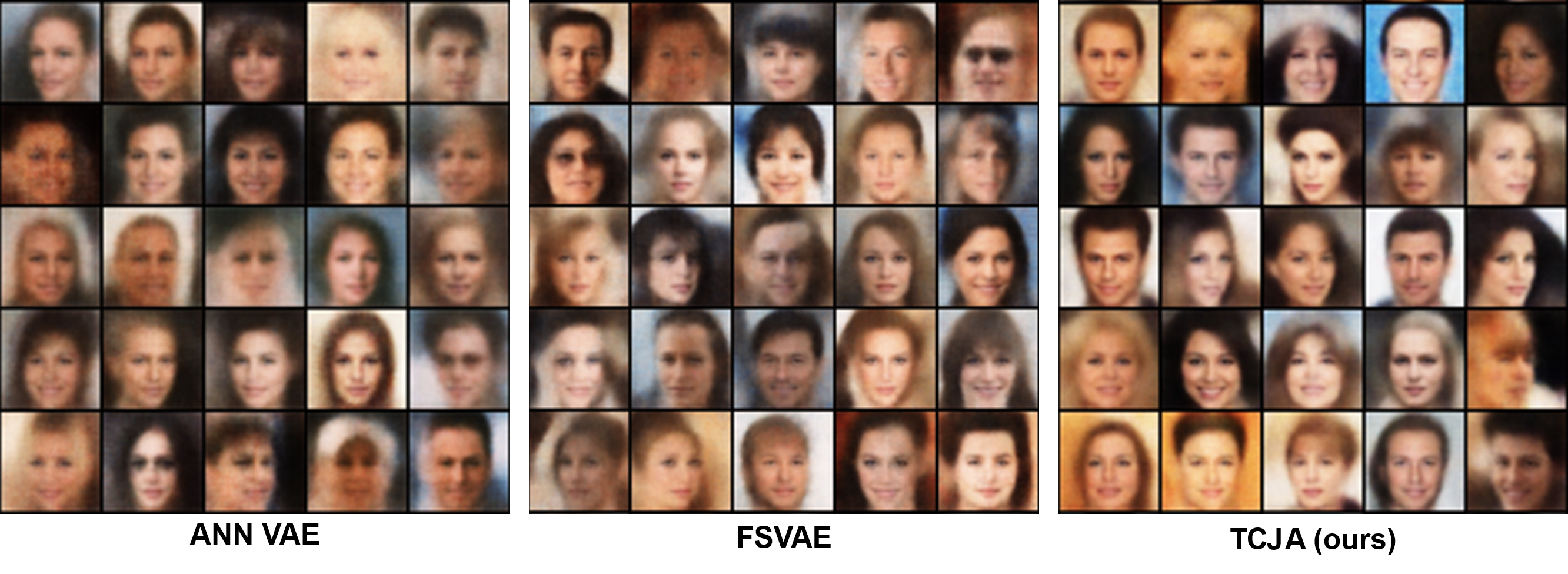}
    \caption{ Generated images of ANN VAE, FSVAE, and ours TCJA on CelebA dataset.}
    \label{qxr3}
\end{figure}

We employ the AdamW optimizer \cite{loshchilov2017decoupled} for image generating tasks, which trains 300 epochs at 0.001 learning rate and 0.001 weight decay. The batch size is set to 256. Moreover, the time step is set to 16. The performance of the TCJA image decode is compared with some SOTA models in Tab. \ref{qxr_tab}. Because this method can make full use of the powerful temporal information, the Inception Score (IS) shows SOTA results compared to the original FSVAE \cite{kamata2022fully} and the same structure ANN. And our TCJA image decoding outperforms better on all metrics for CIFAR10 datasets. Moreover, results on CelebA and MNIST are further visualized in Fig. \ref{qxr2} - \ref{qxr3}, which demonstrates that our generated images are visually better than the previous method \cite{kamata2022fully}. {In addition to the primary evaluations, our model was also compared with other methods based on the Spiking VAEs, such as image decoding based on temporal attention (TAID)~\cite{TAID} and Efficient Spiking VAE (ESVAE)~\cite{zhan2023esvae}, which have been proposed recently. Despite these methods being specifically designed for low-level image reconstruction tasks, our approach remains competitive. It demonstrates robust performance, reflecting its adaptability and effectiveness in comparison to these specialized models.}

\begin{table}[h]
  \centering
  \caption{Comparison with original SNN's work on image generation for each dataset.}
    \begin{tabular}{ccccc}
    \toprule[2pt]
    {Dataset} & {Method} & {IS $\uparrow$} & {FID $\downarrow$}&{FAD $\downarrow$} \\
    \midrule
    {MNIST} & ANN~\cite{kamata2022fully}\textsuperscript{AAAI-2022}   & 5.95  & 112.5 & 17.09 \\
          & FSVAE~\cite{kamata2022fully}\textsuperscript{AAAI-2022} & 6.21  & \textbf{97.06} & 35.54 \\
          & {ESVAE}~\cite{zhan2023esvae} & 5.602  & 117.8 & \textbf{10.99} \\
          & \textbf{This work} & \textbf{6.45} & 100.8 & 19.39 \\
    \midrule
   {Fashion-MNIST} & ANN\cite{kamata2022fully}\textsuperscript{AAAI-2022}   & 4.25  & 123.7 & 18.08 \\
          & FSVAE\cite{kamata2022fully}\textsuperscript{AAAI-2022} & 4.55  & \textbf{90.12} & 15.75 \\
          & {ESVAE}~\cite{zhan2023esvae} & \textbf{6.23}  & 125.3 & \textbf{11.13} \\
          & \textbf{This work} & 5.61 &    93.41   & 12.46 \\
    \midrule
    {CIFAR10} & ANN~\cite{kamata2022fully}\textsuperscript{AAAI-2022}   & 2.59  & 229.6 & 196.9 \\
          & FSVAE~\cite{kamata2022fully}\textsuperscript{AAAI-2022} & 2.94  & 175.5 & 133.9 \\
          & {TAID}~\cite{TAID}\textsuperscript{ICLR-2023}  & 3.53  & 171.1 & 120.5 \\
          & {ESVAE}~\cite{zhan2023esvae} & \textbf{3.76}  & \textbf{127.0} & \textbf{14.7} \\
          & \textbf{This work} & 3.73 &   170.1    & 100.4 \\
    \midrule
    {CelebA} & ANN~\cite{kamata2022fully}\textsuperscript{AAAI-2022}   & 3.23  &  92.53 & 156.9 \\
          & FSVAE~\cite{kamata2022fully}\textsuperscript{AAAI-2022} & 3.69  & 101.6 &  112.9 \\
          & {TAID}~\cite{TAID}\textsuperscript{ICLR-2023}  & \textbf{4.31}  & 99.5 & 105.3 \\
          & {ESVAE}~\cite{zhan2023esvae} & 3.87  & \textbf{85.3} & \textbf{51.9} \\
          & \textbf{This work}&   3.84    &  100.6&  119.9\\
    \bottomrule[2pt]
    \end{tabular}
  \label{qxr_tab}
\end{table}
\subsection{Ablation Study}
\label{sec: asd}
\begin{figure}
     \centering
          \includegraphics[width=0.85\linewidth]{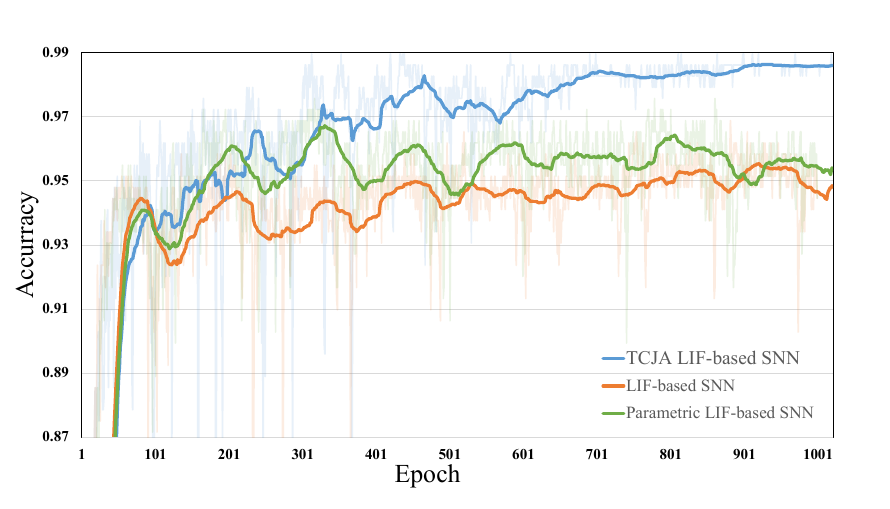}
          \caption{Convergence of compared SNN methods on DVS128 Gesture.}
          \label{fig:convergence}
\end{figure}

\begin{figure}
     \centering
          \includegraphics[width=0.88\linewidth]{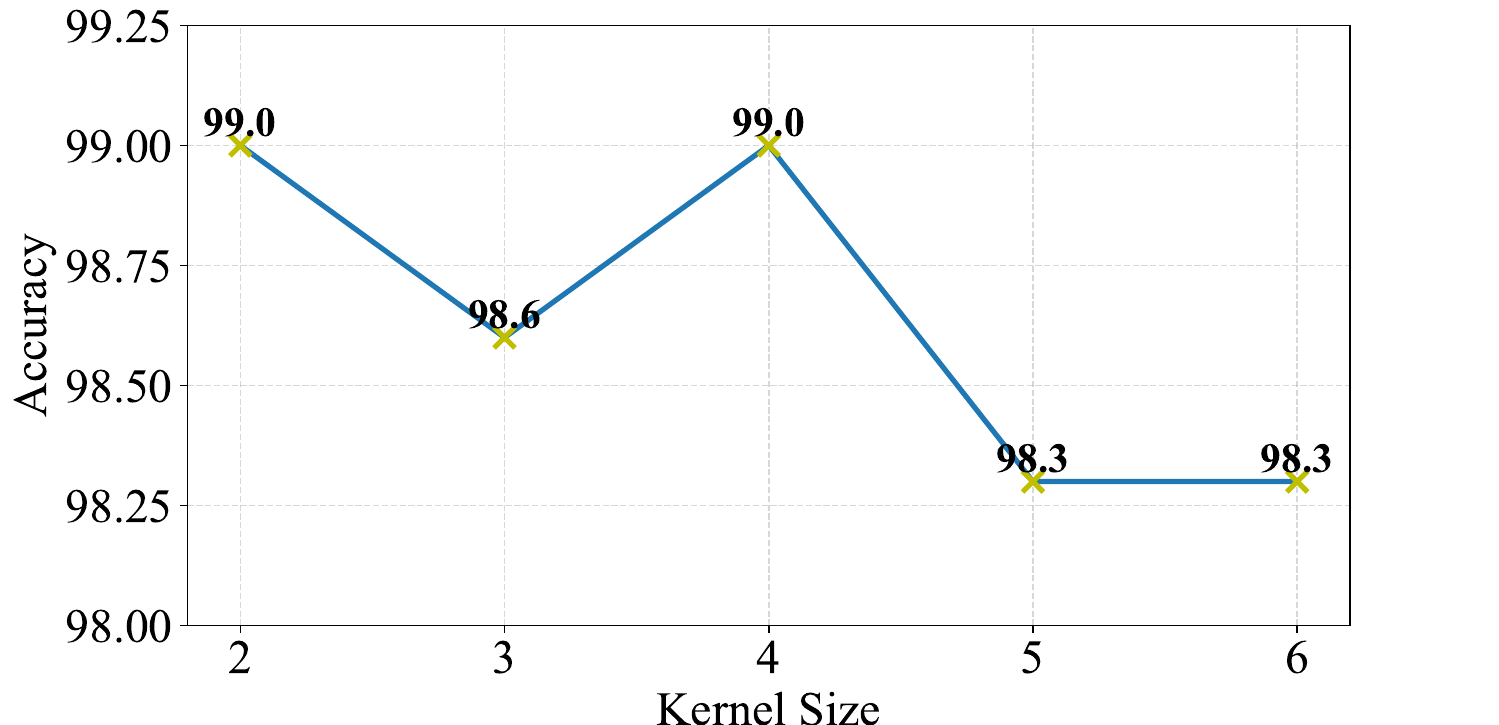}
          \caption{Variation in test accuracy on DVS128 Gesture dataset as kernel size increases.}
          \label{fig:kernel_size}
\end{figure}

\begin{table}[t]
\centering
\caption{Accuracy of different blocks.}
   \begin{adjustbox}{max width=\linewidth}
   \begin{tabular}{lccc}
   \toprule 
   {{Block}} & {CIFAR10-DVS} & {N-Caltech 101} & { DVS128} \\
   \midrule
   TLA & 79.7 & 78.3 & 97.9 \\
   CLA & 80.5 & 78.4 & 98.6 \\
   TCJA & 80.7 & 78.5 & 99.0 \\
   \bottomrule
   \end{tabular}
\label{tab_ablation}
\end{adjustbox}
\end{table}
To thoroughly examine the impact of the TLA and CLA modules, we conducted a series of ablation studies. The results, as presented in Tab. \ref{tab_ablation}, indicate that the CLA module plays a crucial role in enhancing performance. This can be attributed to the fact that, in most SNN designs, the number of simulation time steps is significantly fewer than the number of channels. Consequently, the CLA module is able to extract additional relevant features compared to the TLA module. Furthermore, it is worth noting that the TCJA module consistently outperformed other models across all tested datasets. This outcome underscores the effectiveness of the CCF layer incorporated within the TCJA module, further reinforcing its potential for achieving superior performance.
  \begin{figure*}
    \centering
    \includegraphics[width=0.70\textwidth]{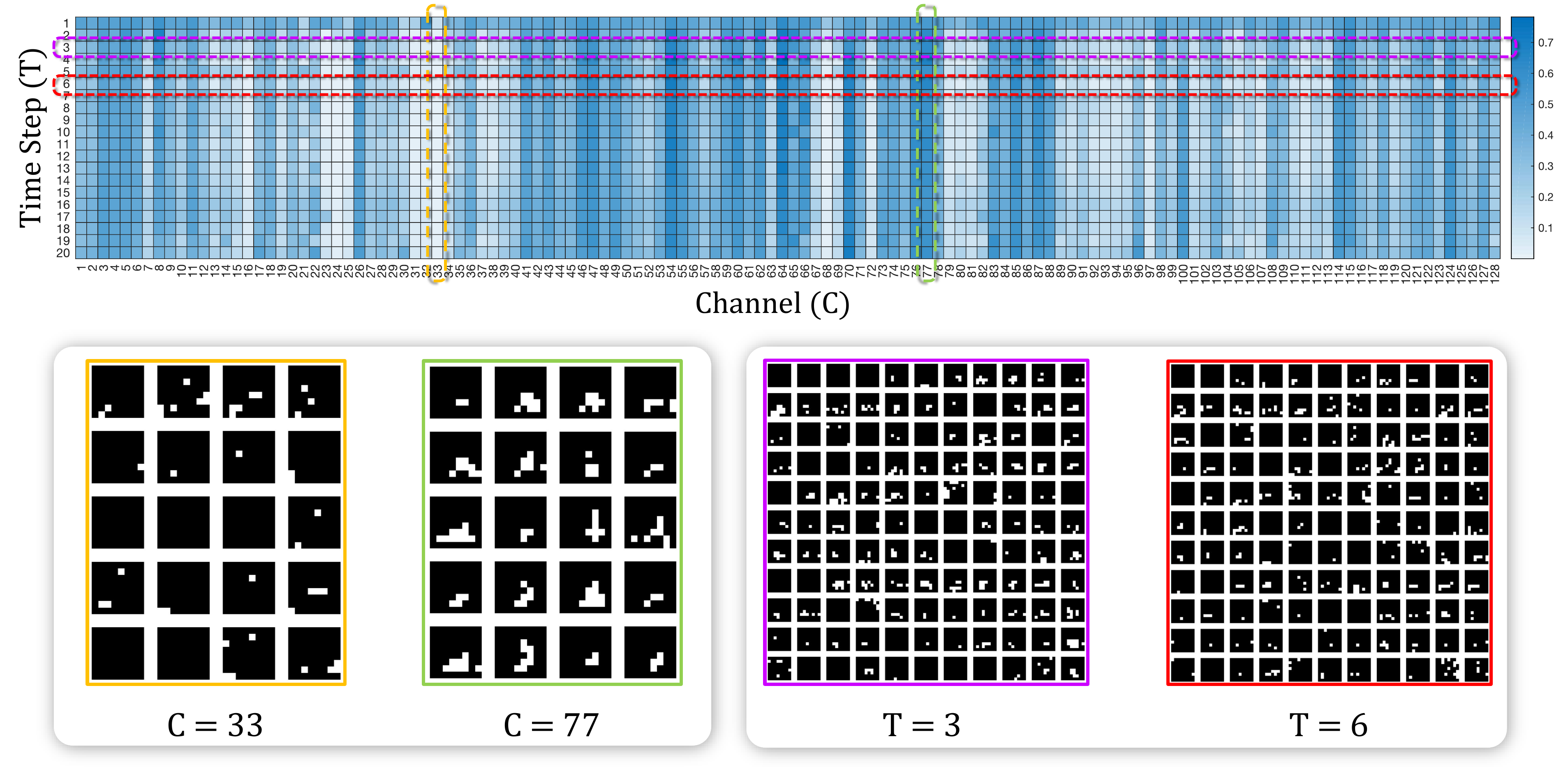}
    \caption{Attention distribution between time step and channel. The top row is the weight from the first TCJA module in TCJA-SNN working with the DVS128 Gesture dataset. We select sparse and dense attention frames in both temporal-wise ($T=3,6$) and channel-wise ($C=33,77$) in the bottom row.}
    \label{fig:attention_distri}
\end{figure*}
\subsection{Discussion}
\subsubsection{Kernel Size}
We initially investigate the kernel size in the TCJA module. Intuitively, when the size of the kernel rises, the receptive field of the local attention mechanism will also expand, which may aid in enhancing the performance of TCJA-SNN. However, the experimental results in Fig. \ref{fig:kernel_size} overturn this conjecture. As the size of the kernel rises, the performance of the model waves. When the kernel size is more than 4, there is a perceptible decrease in overall performance. One reasonable explanation is that a frame mainly correlates with its nearby frames, and an excessively large receptive field may lead to undesired noise.

\subsubsection{Multiplication vs. Addition}
To verify the effectiveness of our proposed CCF mechanism, we devise a variant method that substitutes addition for multiplication of $\mathcal{T}_{i,j}$ and $\mathcal{C}_{i,j}$ in the Eq. ~\ref{equ_ccf}. The results are shown in Tab.~\ref{tab:mva}.
\begin{table}[htbp]
\centering
\caption{Test accuracy on three datasets with different CCF operations.}
\begin{tabular}{lccc}
\toprule 
{{\textbf{Type}}} & {\textbf{CIFAR10-DVS}} & {\textbf{N-Caltech 101}} & {\textbf{DVS128}} \\
\midrule
Addition & 80.7 & 78.0 & 98.2 \\
Multiplication & 80.7 & 78.5 & 99.0 \\
\bottomrule
\label{tab:mva}
\end{tabular}
\end{table}

As we observed, the addition operation achieves good performance, nevertheless, when compared to the multiplication operation, the final calculation result lacks the cross term, which prevents a robust construction of the correlation between frames; therefore, it is inferior.

\subsubsection{Convergence}We also empirically demonstrate the convergence of our proposed method, as shown in Fig. \ref{fig:convergence}. {Specifically, Fig. \ref{fig:convergence} illustrates the performance trend of vanilla LIF-SNN, Parametric LIF-based SNN~\cite{fang2021incorporating} without TCJA block and our proposed TCJA-SNN for 1000 epochs.} As the training epoch increases, the performance trend of our proposed method becomes more stable and converges to a higher level. 
Moreover, the TCJA-SNN can achieve the SOTA performance when only training about 260 epochs, which demonstrates the efficacy of the proposed TCJA.

\subsubsection{Complexity Analysis}
Finally, we try to analyze the time and space complexity of TLA and CLA. For TLA, it can be concluded from the Eq.~\ref{equ_time_attention} that the time complexity of obtaining each element of $\mathcal{T}$ is $\mathcal{O} (CK)$ ($K$: Kernel size.). Consequently, the time complexity of the whole TLA mechanism is $\mathcal{O} (TC^2K)$. Moreover, the space complexity is composed of the parameters and the memory occupied by variables. On the one hand, for the parameters, C-channel 1-D convolution is performed on each row of $\mathcal{Z}$, so the total amount of parameters required is $C*C*K$, on the other hand, for variables, in the whole process, we only need to maintain a matrix of dimension $C\times T$. In conclusion, the space complexity is $\mathcal{O} (C^2K+CT)$. Similarly, for CLA, the time complexity is $\mathcal{O} (T^2CK)$ and the space complexity is $\mathcal{O} (T^2K+CT)$.

\subsubsection{Theoretical Analysis on Receptive Field}
\label{sec:theoretical}
The global receptive field stands as a fundamental feature of our innovative TCJA approach. Our approach surpasses the limitations of dense layers by utilizing fewer parameters to achieve a comprehensive global receptive field. Moreover, it surpasses the capabilities of employing 2D convolutions alone by effectively obtaining a larger receptive field. In order to provide a deeper comprehension of the salient aspects of our proposed method, we present the following theoretical analysis concerning the specific region where the network perceives and processes information throughout the training phase, known as the receptive field.

\textbf{Lemma 1.} \textit{(\textbf{Cross-Correlation Scope (CCS) of 1-D convolution}) For an input feature map $I\in \mathbb{R}^{C \times T}$, if the size of the 1-D convolution kernel is defined as $k$, then its CCS can be described as $P \in \mathbb{R}^{k \times T}$, where the $T$ involves the information along the second dimension of $I$. }

\textbf{Lemma 2.} \textit{(\textbf{CCS of two orthogonal 1-D convolution}) For an input feature map $I\in \mathbb{R}^{C \times T}$, the dot multiplication of two orthogonal 1-D convolutions performed on $I$ is equivalent to expanding the CCS into a cross shape, \textit{i.e.}, its CCS can be described by two cross-overlaid matrices $P \ast Q$ (see e.g., the colored area of $\mathcal{F}$ in Fig. \ref{fig_TCJA_demo}), where $P \in \mathbb{R}^{k_1 \times T}$, $Q \in \mathbb{R}^{k_2 \times C}$, and ${k}_{1}$ and ${k}_{2}$ are the sizes of the two convolution kernels, respectively.} 

 Referring to Eq. ~\ref{equ_ccf}, Lemma 1, and Lemma 2, we can obtain the following corollary:

\textbf{Corollary.} \textit{Based on the broad CCS obtained by TCJA, there exists information flow among $\mathcal{T}$ and $\mathcal{C}$, cooperatively considering the temporal and channel correlation, which is also clued in Eq. ~\ref{equ_ccf}.}

Recalling Eq. ~\ref{equ_time_attention} and Eq. ~\ref{equ_channel_attention}, through two 1-D convolutions along different dimensions, we construct two CCS in a vertical relationship, which are stored in $\mathcal{T}$ and $\mathcal{C}$. In particular, TCJA is to construct a CCS, which can perceive a larger area while realizing feature interaction in different directions. This cross-receptive field is able to abolish the limitations 
caused by the monotonic dimension, thus bringing performance improvements to the network. {As a corollary, when the kernel sizes of the two dimensions are the same, we can obtain a square cross-shaped receptive field similar to that of conventional 2-D convolution, which is a effective scheme in 2-D convolution.}

\subsubsection{Attention Visualization}
\label{sec: DAV}
To make the attention mechanism easier to understand, we finally visualize the output of the first TCJA module in TCJA-SNN working with the DVS128 Gesture dataset, which can be seen in Fig. \ref{fig:attention_distri}. Changes in attention weights are primarily accumulated among channels, verifying further the substantial role performed by the CLA in the TCJA module. To embody the attention weights, we extract some temporal-wise and channel-wise frames. The difference in firing patterns in the channel dimension is more significant than that in the temporal dimension. 

\subsubsection{Energy Consumption Analysis}
{Compared to the ANNs, SNNs consumes less energy due to its sparser firing and poorer processing accuracy.
Owing to the binary spikes, each operation in SNNs consists of a single floating-point (FP) addition. In ANNs, on the other hand, each operation computes a dot product as a multiply-accumulate (MAC) calculation consisting of one floating-point (FP) multiplication and one FP addition. }Consequently, SNNs use less energy than ANNs in the same network design. This discrepancy can also be validated in 45nm CMOS technology, where the energy cost of each SNN operation is 5.1$\times$ lower than that of each 32-bit ANN MAC operation (0.9pJ vs. 4.6pJ) \cite{horowitz20141}, allowing us to examine the energy consumption of each network architecture.

We assess the energy consumption in the network for classifying the DVS128 dataset with both ANN and SNN. first, we assess the spiking rate, floating-point operations per second (FLOPS) of each layer, the result is shown in Tab. \ref{tab:energy}.
\begin{table*}
\centering

\caption{The spiking rate, FLOPS, and SNN single operation energy cost of each layer in the network for classifying the DVS128 dataset, where Conv$x$ denotes $x-th$ 2-D convolutional layer, Att$y$ denotes $y-th$ TCJA module, and FC$z$ represents $z-th$ fully-connected layer of the network. Notice that the first 2-D convolutional layer is an encoding layer to transform the analog input into spikes.}
\label{tab:energy}
\begin{tabular}{lccccccccc}
\toprule
\textbf{Layer}  & \textbf{Encoding} & \textbf{Conv1}   & \textbf{Conv2}   & \textbf{Conv3}  & \textbf{Conv4}  & \textbf{Att1}  & \textbf{Att2}  & \textbf{FC1}    & \textbf{FC2}    \\\midrule
\textbf{Spiking Rate}                                                            & -        & 2.27\%  & 2.50\%  & 4.74\% & 8.09\% & -     & -     & 27.80\% & 9.01\% \\
\textbf{FLOPS}                                                                    & 37.74M   & 603.98M & 151.00M & 37.74M & 9.44M  & 1.31M & 1.31M & 1.05M  & 0.06M  \\
\begin{tabular}[c]{@{}l@{}}\textbf{Energy Cost in SNN}\\ \textbf{(Single Operation)}\end{tabular} & 4.6pJ    & 0.9pJ   & 0.9pJ   & 0.9pJ  & 0.9pJ  & 4.6pJ & 4.6pJ & 0.9pJ  & 0.9pJ \\ \bottomrule
\end{tabular}
\end{table*}
For ANN, we are able to calculate the energy consumption by $FLOPS \times MAC\;energy\;cost$; for SNN, the energy cost should be quantified by $FLOPS \times SNN\; operation\;energy\;cost \times spiking\;rate$. The final power consumption calculation results are $1.90 \times 10^{-3}$J (TCJA-SNN) and $10.00 \times 10^{-3}$J (ANN), where our TCJA-SNN costs $5.26\times$ lower energy consumption compared to its ANN version.

\begin{figure}[htbp]
    \centering
    \includegraphics[width=\linewidth]{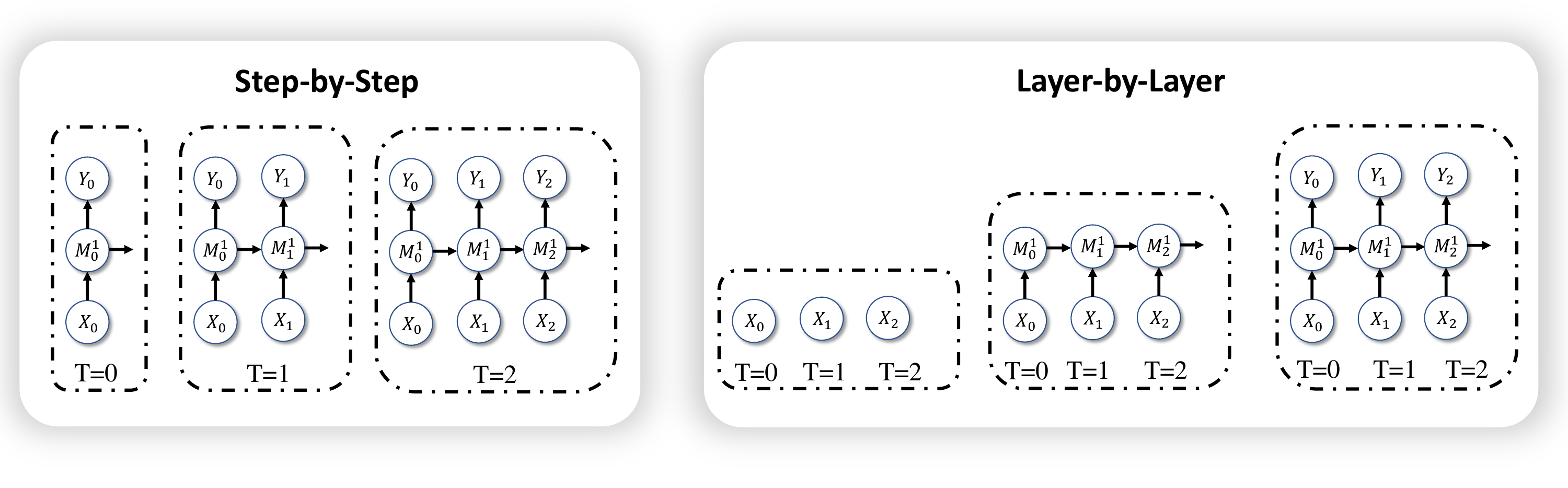}
    \caption{step-by-step propagation pattern and layer-by-layer propagation pattern. The $X_{m}$ denotes the input in $m-th$ time-step, and the $M^{i}_{j}$ represents the $i-th$ middle layer in $j-th$ time-step. The $Y_{n}$ shows the output in $n-th$ time-step.}
    \label{fig:prop_pattern}
\end{figure}
\subsubsection{Propagation Pattern}

{The forward propagation process in SNNs encompasses both temporal and spatial domains. Intuitively, the computation graph for SNN forward propagation can be conceptualized as a sequential, step-by-step pattern. This pattern is depicted in Fig. \ref{fig:prop_pattern}. In this context, ``  step-by-step" refers to the process wherein the network's output at the initial timestep is evaluated, along with updates to the hidden states of the spiking neurons. Following this, subsequent timesteps are evaluated in a similar manner, maintaining this sequential progression.} Besides, the layer-by-layer pattern is also extensively used, which entails performing a spatial forward propagation procedure in which we calculate the output of the first layer at all timesteps as the input of the second layer, then retrieve the output of the last layer at all timesteps. Unlike parallel computing environments like GPU, where the layer-by-layer pattern is preferred, neuromorphic devices operate more like a step-by-step pattern. It can be proved that the output of the network in the two patterns is mathematically equivalent.

Although we train the network in TCJA using a layer-by-layer pattern, the convolutional structure of the network still benefits it when applied in a step-by-step manner. In terms of temporal attention, the TLA module only needs to compute a few adjacent timesteps because of its convolutional nature, in contrast to mechanisms like squeeze and excitation that require full information at all timesteps. Previous discussion reveals that TCJA reaches its peak performance when the convolutional kernel size is set to 2. Under this circumstance, TLA only needs to buffer one timestep while propagating step-by-step.
\section{Conclusion}
\label{sec: con}
In this paper, we propose the TCJA mechanism, which innovatively recalibrates temporal and channel information in SNN. Specifically, instead of utilizing a generic fully connected network, we use 1-D convolution to build the correlation between frames, reducing the computation and improving model performance. Moreover, we propose a CCF mechanism to realize joint feature interaction between temporal and channel information. {Experiments verify the effectiveness of our method with SOTA results on four datasets, \textit{i.e.}, CIFAR10-DVS (83.3\%), N-Caltech101 (82.5\%), DVS128 (99.0\%), Fashion-MNIST (94.8\%), CIFAR10 (95.9\%) and CIFAR100 (78.3\%). In addition to its outstanding performance in classification tasks, TCJA-SNN also exhibits a competitive performance in image generation tasks.} To the best of our knowledge, this study represents the pioneering application of the SNN-attention mechanism to both high-level classification and low-level generation tasks. Remarkably, our approach has achieved state-of-the-art performance in both domains, thus making a significant advancement in the field. However, the insertion of TCJA still resulted in a relatively sizable boost in the number of parameters.
In future work, we believe this method can easily be integrated into the neuromorphic chip for the hardware-friendly 1-D convolution operation and the binary spiking network structure.
\section*{Acknowledgments}
We would like to express our gratitude to Xuerui Qiu for providing insightful comments on our work and for his efforts in exploring the application of TCJA for generation tasks. This research is supported by National Natural Science Foundation of China (No. 12271083, 62106038, 12171072), and National Key Research and Development Program of China (No. 2020YFA0714001).

\bibliography{reference.bib}

\begin{thebibliography}{10}
\providecommand{\url}[1]{#1}
\csname url@samestyle\endcsname
\providecommand{\newblock}{\relax}
\providecommand{\bibinfo}[2]{#2}
\providecommand{\BIBentrySTDinterwordspacing}{\spaceskip=0pt\relax}
\providecommand{\BIBentryALTinterwordstretchfactor}{4}
\providecommand{\BIBentryALTinterwordspacing}{\spaceskip=\fontdimen2\font plus
\BIBentryALTinterwordstretchfactor\fontdimen3\font minus \fontdimen4\font\relax}
\providecommand{\BIBforeignlanguage}[2]{{%
\expandafter\ifx\csname l@#1\endcsname\relax
\typeout{** WARNING: IEEEtran.bst: No hyphenation pattern has been}%
\typeout{** loaded for the language `#1'. Using the pattern for}%
\typeout{** the default language instead.}%
\else
\language=\csname l@#1\endcsname
\fi
#2}}
\providecommand{\BIBdecl}{\relax}
\BIBdecl

\bibitem{roy2019towards}
K.~Roy, A.~Jaiswal, and P.~Panda, ``Towards spike-based machine intelligence with neuromorphic computing,'' \emph{Nature}, vol. 575, no. 7784, pp. 607--617, 2019.

\bibitem{stromatias2015robustness}
E.~Stromatias, D.~Neil, M.~Pfeiffer, F.~Galluppi, S.~B. Furber, and S.-C. Liu, ``{Robustness of spiking Deep Belief Networks to noise and reduced bit precision of neuro-inspired hardware platforms},'' \emph{Frontiers in Neuroscience}, vol.~9, p. 222, 2015.

\bibitem{zhang2021rectified}
M.~Zhang, J.~Wang, J.~Wu, A.~Belatreche, B.~Amornpaisannon, Z.~Zhang, V.~P.~K. Miriyala, H.~Qu, Y.~Chua, T.~E. Carlson \emph{et~al.}, ``Rectified linear postsynaptic potential function for backpropagation in deep spiking neural networks,'' \emph{IEEE Transactions on Neural Networks and Learning Systems}, vol.~33, no.~5, pp. 1947--1958, 2021.

\bibitem{bohte2000spikeprop}
S.~M. Bohte, J.~N. Kok, and J.~A. La~Poutr{\'e}, ``Spikeprop: backpropagation for networks of spiking neurons.'' in \emph{ESANN}, vol.~48.\hskip 1em plus 0.5em minus 0.4em\relax Bruges, 2000, pp. 419--424.

\bibitem{wu2021progressive}
J.~Wu, C.~Xu, X.~Han, D.~Zhou, M.~Zhang, H.~Li, and K.~C. Tan, ``Progressive tandem learning for pattern recognition with deep spiking neural networks,'' \emph{IEEE Transactions on Pattern Analysis and Machine Intelligence}, vol.~44, no.~11, pp. 7824--7840, 2021.

\bibitem{dampfhoffer2023backpropagation}
M.~Dampfhoffer, T.~Mesquida, A.~Valentian, and L.~Anghel, ``Backpropagation-based learning techniques for deep spiking neural networks: A survey,'' \emph{IEEE Transactions on Neural Networks and Learning Systems}, 2023.

\bibitem{lee2016training}
J.~H. Lee, T.~Delbruck, and M.~Pfeiffer, ``Training deep spiking neural networks using backpropagation,'' \emph{Frontiers in neuroscience}, vol.~10, p. 508, 2016.

\bibitem{luo2022supervised}
X.~Luo, H.~Qu, Y.~Wang, Z.~Yi, J.~Zhang, and M.~Zhang, ``Supervised learning in multilayer spiking neural networks with spike temporal error backpropagation,'' \emph{IEEE Transactions on Neural Networks and Learning Systems}, 2022.

\bibitem{cao2015spiking}
Y.~Cao, Y.~Chen, and D.~Khosla, ``Spiking deep convolutional neural networks for energy-efficient object recognition,'' \emph{International Journal of Computer Vision}, vol. 113, pp. 54--66, 2015.

\bibitem{zhang2023self}
T.~Zhang, Q.~Wang, and B.~Xu, ``Self-lateral propagation elevates synaptic modifications in spiking neural networks for the efficient spatial and temporal classification,'' \emph{IEEE Transactions on Neural Networks and Learning Systems}, 2023.

\bibitem{zheng2021going}
H.~Zheng, Y.~Wu, L.~Deng, Y.~Hu, and G.~Li, ``{Going Deeper With Directly-Trained Larger Spiking Neural Networks},'' in \emph{Proceedings of the AAAI Conference on Artificial Intelligence (AAAI)}, 2021, pp. 11\,062--11\,070.

\bibitem{hu2018spiking}
Y.~Hu, H.~Tang, and G.~Pan, ``{Spiking Deep Residual Networks},'' \emph{IEEE Transactions on Neural Networks and Learning Systems}, pp. 1--6, 2018.

\bibitem{wu2019direct}
Y.~Wu, L.~Deng, G.~Li, J.~Zhu, Y.~Xie, and L.~Shi, ``{Direct Training for Spiking Neural Networks: Faster, Larger, Better},'' in \emph{Proceedings of the AAAI Conference on Artificial Intelligence (AAAI)}, 2019, pp. 1311--1318.

\bibitem{yao2021temporal}
M.~Yao, H.~Gao, G.~Zhao, D.~Wang, Y.~Lin, Z.~Yang, and G.~Li, ``{Temporal-wise Attention Spiking Neural Networks for Event Streams Classification},'' in \emph{Proceedings of the IEEE/CVF International Conference on Computer Vision (ICCV)}, 2021, pp. 10\,201--10\,210.

\bibitem{hu2018squeeze}
J.~Hu, L.~Shen, and G.~Sun, ``{Squeeze-and-Excitation Networks},'' in \emph{Proceedings of the IEEE Conference on Computer Vision and Pattern Recognition (CVPR)}, 2018.

\bibitem{woo2018cbam}
S.~Woo, J.~Park, J.-Y. Lee, and I.~S. Kweon, ``Cbam: Convolutional block attention module,'' in \emph{Proceedings of the European conference on computer vision (ECCV)}, 2018, pp. 3--19.

\bibitem{Bernert2018Dec}
M.~Bernert and B.~Yvert, ``{An Attention-Based Spiking Neural Network for Unsupervised Spike-Sorting},'' \emph{International Journal of Neural Systems}, vol.~29, no.~8, pp. 1\,850\,059:1--1\,850\,059:19, 2019.

\bibitem{wang2023toward}
Z.~Wang, Y.~Zhang, S.~Lian, X.~Cui, R.~Yan, and H.~Tang, ``Toward high-accuracy and low-latency spiking neural networks with two-stage optimization,'' \emph{IEEE Transactions on Neural Networks and Learning Systems}, 2023.

\bibitem{wu2021tandem}
J.~Wu, Y.~Chua, M.~Zhang, G.~Li, H.~Li, and K.~C. Tan, ``A tandem learning rule for effective training and rapid inference of deep spiking neural networks,'' \emph{IEEE Transactions on Neural Networks and Learning Systems}, 2021.

\bibitem{yang2023lc}
Q.~Yang, M.~Zhang, J.~Wu, K.~C. Tan, and H.~Li, ``Lc-ttfs: Towards lossless network conversion for spiking neural networks with ttfs coding,'' \emph{IEEE Transactions on Cognitive and Developmental Systems}, 2023.

\bibitem{fang2021deep}
W.~Fang, Z.~Yu, Y.~Chen, T.~Huang, T.~Masquelier, and Y.~Tian, ``{Deep Residual Learning in Spiking Neural Networks},'' in \emph{Advances in Neural Information Processing Systems (NeurIPS)}, vol.~34, 2021, pp. 21\,056--21\,069.

\bibitem{jin2022sit}
C.~Jin, R.-J. Zhu, X.~Wu, and L.-J. Deng, ``{SIT: A Bionic and Non-Linear Neuron for Spiking Neural Network},'' \emph{ArXiv preprint arXiv:2203.16117}, 2022.

\bibitem{Neftci2019Nov}
E.~O. Neftci, H.~Mostafa, and F.~Zenke, ``Surrogate gradient learning in spiking neural networks: Bringing the power of gradient-based optimization to spiking neural networks,'' \emph{IEEE Signal Processing Magazine}, vol.~36, no.~6, pp. 51--63, 2019.

\bibitem{wu2018spatio}
Y.~Wu, L.~Deng, G.~Li, J.~Zhu, and L.~Shi, ``{Spatio-Temporal Backpropagation for Training High-performance Spiking Neural Networks},'' \emph{Frontiers in neuroscience}, vol.~12, p. 331, 2018.

\bibitem{rathi2021diet}
N.~Rathi and K.~Roy, ``Diet-snn: A low-latency spiking neural network with direct input encoding and leakage and threshold optimization,'' \emph{IEEE Transactions on Neural Networks and Learning Systems}, 2021.

\bibitem{xie2023event}
X.~Xie, Y.~Chua, G.~Liu, M.~Zhang, G.~Luo, and H.~Tang, ``Event-driven spiking learning algorithm using aggregated labels,'' \emph{IEEE Transactions on Neural Networks and Learning Systems}, 2023.

\bibitem{qiu2023gated}
X.~Qiu, R.-J. Zhu, Y.~Chou, Z.~Wang, L.-j. Deng, and G.~Li, ``Gated attention coding for training high-performance and efficient spiking neural networks,'' \emph{arXiv preprint arXiv:2308.06582}, 2023.

\bibitem{qiu2023vtsnn}
X.-R. Qiu, Z.-R. Wang, Z.~Luan, R.-J. Zhu, M.-L. Zhang, and L.-J. Deng, ``Vtsnn: a virtual temporal spiking neural network,'' \emph{Frontiers in neuroscience}, vol.~17, p. 1091097, 2023.

\bibitem{Eshraghian2021Sep}
J.~K. Eshraghian, M.~Ward, E.~Neftci, X.~Wang, G.~Lenz, G.~Dwivedi, M.~Bennamoun, D.~S. Jeong, and W.~D. Lu, ``{Training Spiking Neural Networks Using Lessons From Deep Learning},'' \emph{ArXiv}, 2021.

\bibitem{fang2021incorporating}
W.~Fang, Z.~Yu, Y.~Chen, T.~Masquelier, T.~Huang, and Y.~Tian, ``{Incorporating Learnable Membrane Time Constant To Enhance Learning of Spiking Neural Networks},'' in \emph{Proceedings of the IEEE/CVF International Conference on Computer Vision (ICCV)}, 2021, pp. 2661--2671.

\bibitem{bellec2018long}
G.~Bellec, D.~Salaj, A.~Subramoney, R.~Legenstein, and W.~Maass, ``{Long short-term memory and learning-to-learn in networks of spiking neurons},'' in \emph{Advances in Neural Information Processing Systems (NeurIPS)}, vol.~31, 2018.

\bibitem{mante2008functional}
V.~Mante, V.~Bonin, and M.~Carandini, ``Functional mechanisms shaping lateral geniculate responses to artificial and natural stimuli,'' \emph{Neuron}, vol.~58, no.~4, pp. 625--638, 2008.

\bibitem{izhikevich2003simple}
E.~M. Izhikevich, ``Simple model of spiking neurons,'' \emph{IEEE Transactions on neural networks}, vol.~14, no.~6, pp. 1569--1572, 2003.

\bibitem{gerstner2002spiking}
W.~Gerstner and W.~M. Kistler, \emph{Spiking neuron models: Single neurons, populations, plasticity}.\hskip 1em plus 0.5em minus 0.4em\relax Cambridge university press, 2002.

\bibitem{lapicque1907louis}
L.~Lapique, ``Recherches quantitatives sur l'excitation electrique des nerfs traitee comme une polarization,'' \emph{Journal of Physiology and Pathology}, vol.~9, pp. 620--635, 1907.

\bibitem{deng2021temporal}
S.~Deng, Y.~Li, S.~Zhang, and S.~Gu, ``{Temporal Efficient Training of Spiking Neural Network via Gradient Re-weighting},'' in \emph{International Conference on Learning Representations (ICLR)}, 2021.

\bibitem{simonyan2014very}
K.~Simonyan and A.~Zisserman, ``Very deep convolutional networks for large-scale image recognition,'' \emph{arXiv preprint arXiv:1409.1556}, 2014.

\bibitem{luo2016understanding}
W.~Luo, Y.~Li, R.~Urtasun, and R.~Zemel, ``Understanding the effective receptive field in deep convolutional neural networks,'' \emph{Advances in neural information processing systems}, vol.~29, 2016.

\bibitem{li2017cifar10dvs}
H.~Li, H.~Liu, X.~Ji, G.~Li, and L.~Shi, ``{CIFAR10-DVS: An Event-Stream Dataset for Object Classification},'' \emph{Frontiers in Neuroscience}, vol.~11, p. 309, 2017.

\bibitem{li2021differentiable}
Y.~Li, Y.~Guo, S.~Zhang, S.~Deng, Y.~Hai, and S.~Gu, ``{Differentiable Spike: Rethinking Gradient-Descent for Training Spiking Neural Networks},'' in \emph{Advances in Neural Information Processing Systems (NeurIPS)}, vol.~34, 2021, pp. 23\,426--23\,439.

\bibitem{li2022neuromorphic}
Y.~Li, Y.~Kim, H.~Park, T.~Geller, and P.~Panda, ``{Neuromorphic Data Augmentation for Training Spiking Neural Networks},'' \emph{ArXiv preprint arXiv:2203.06145}, 2022.

\bibitem{meng2022training}
Q.~Meng, M.~Xiao, S.~Yan, Y.~Wang, Z.~Lin, and Z.-Q. Luo, ``{Training High-Performance Low-Latency Spiking Neural Networks by Differentiation on Spike Representation},'' \emph{ArXiv preprint arXiv:2205.00459}, 2022.

\bibitem{orchard2015nmnist}
G.~Orchard, A.~Jayawant, G.~K. Cohen, and N.~Thakor, ``{Converting Static Image Datasets to Spiking Neuromorphic Datasets Using Saccades},'' \emph{Frontiers in Neuroscience}, vol.~9, p. 437, 2015.

\bibitem{1384978}
L.~Fei-Fei, R.~Fergus, and P.~Perona, ``{Learning Generative Visual Models from Few Training Examples: An Incremental Bayesian Approach Tested on 101 Object Categories},'' in \emph{2004 Conference on Computer Vision and Pattern Recognition (CVPR) Workshop}, 2004, pp. 178--178.

\bibitem{amir2017dvsg}
A.~Amir, B.~Taba, D.~Berg, and et~al., ``{A Low Power, Fully Event-Based Gesture Recognition System},'' in \emph{Proceedings of the IEEE/CVF Conference on Computer Vision and Pattern Recognition (CVPR)}, 2017.

\bibitem{xiao2017fashionmnist}
H.~Xiao, K.~Rasul, and R.~Vollgraf, ``{Fashion-mnist: a Novel Image Dataset for Benchmarking Machine Learning Algorithms},'' \emph{ArXiv preprint arXiv:1708.07747}, 2017.

\bibitem{krizhevsky2009cifar10}
A.~Krizhevsky, G.~Hinton \emph{et~al.}, ``{Learning Multiple Layers of Features from Tiny Images},'' 2009.

\bibitem{kugele2020efficient}
A.~Kugele, T.~Pfeil, M.~Pfeiffer, and E.~Chicca, ``{Efficient Processing of Spatio-temporal Data Streams with Spiking Neural Networks},'' \emph{Frontiers in Neuroscience}, vol.~14, p. 439, 2020.

\bibitem{zhang2018mixup}
H.~Zhang, M.~Cisse, Y.~N. Dauphin, and D.~Lopez-Paz, ``{mixup: Beyond Empirical Risk Minimization},'' in \emph{International Conference on Learning Representations (ICLR)}, 2018.

\bibitem{he2015delving}
K.~He, X.~Zhang, S.~Ren, and J.~Sun, ``Delving deep into rectifiers: Surpassing human-level performance on imagenet classification,'' in \emph{Proceedings of the IEEE international conference on computer vision}, 2015, pp. 1026--1034.

\bibitem{srivastava2014dropout}
N.~Srivastava, G.~Hinton, A.~Krizhevsky, I.~Sutskever, and R.~Salakhutdinov, ``Dropout: A simple way to prevent neural networks from overfitting,'' \emph{The Journal of Machine Learning Research}, vol.~15, no.~1, pp. 1929--1958, 2014.

\bibitem{hu2021advancing}
Y.~Hu, Y.~Wu, L.~Deng, and G.~Li, ``Advancing deep residual learning by solving the crux of degradation in spiking neural networks,'' \emph{arXiv preprint arXiv:2201.07209}, 2021.

\bibitem{SpikingJelly}
W.~Fang, Y.~Chen, J.~Ding, D.~Chen, Z.~Yu, H.~Zhou, Y.~Tian, and other contributors, ``Spikingjelly,'' \url{https://github.com/fangwei123456/spikingjelly}, 2020, accessed: 2022-05-04.

\bibitem{paszke2019pytorch}
A.~Paszke, S.~Gross, F.~Massa, A.~Lerer, J.~Bradbury, G.~Chanan, T.~Killeen, Z.~Lin, N.~Gimelshein, L.~Antiga, A.~Desmaison, A.~Kopf, E.~Yang, Z.~DeVito, M.~Raison, A.~Tejani, S.~Chilamkurthy, B.~Steiner, L.~Fang, J.~Bai, and S.~Chintala, ``{PyTorch: An Imperative Style, High-Performance Deep Learning Library},'' in \emph{Advances in Neural Information Processing Systems (NeurIPS)}, vol.~32, 2019.

\bibitem{wu2021liaf}
Z.~Wu, H.~Zhang, Y.~Lin, G.~Li, M.~Wang, and Y.~Tang, ``{LIAF-Net: Leaky Integrate and Analog Fire Network for Lightweight and Efficient Spatiotemporal Information Processing},'' \emph{IEEE Transactions on Neural Networks and Learning Systems}, pp. 1--14, 2021.

\bibitem{yaoglif}
X.~Yao, F.~Li, Z.~Mo, and J.~Cheng, ``Glif: A unified gated leaky integrate-and-fire neuron for spiking neural networks,'' in \emph{Advances in Neural Information Processing Systems (NeurIPS)}, 2022.

\bibitem{shrestha2018slayer}
S.~B. Shrestha and G.~Orchard, ``{SLAYER: Spike Layer Error Reassignment in Time},'' in \emph{Advances in Neural Information Processing Systems (NeurIPS)}, vol.~31, 2018.

\bibitem{sironi2018hats}
A.~Sironi, M.~Brambilla, N.~Bourdis, X.~Lagorce, and R.~Benosman, ``{HATS: Histograms of Averaged Time Surfaces for Robust Event-Based Object Classification},'' in \emph{Proceedings of the IEEE/CVF Conference on Computer Vision and Pattern Recognition (CVPR)}, 2018, pp. 1731--1740.

\bibitem{ramesh2019dart}
B.~Ramesh, H.~Yang, G.~Orchard, N.~A. Le~Thi, S.~Zhang, and C.~Xiang, ``{DART: Distribution Aware Retinal Transform for Event-Based Cameras},'' \emph{IEEE Transactions on Pattern Analysis and Machine Intelligence}, vol.~42, no.~11, pp. 2767--2780, 2019.

\bibitem{kaiser2020synaptic}
J.~Kaiser, H.~Mostafa, and E.~Neftci, ``{Synaptic Plasticity Dynamics for Deep Continuous Local Learning (DECOLLE)},'' \emph{Frontiers in Neuroscience}, vol.~14, p. 424, 2020.

\bibitem{kim2021optimizing}
Y.~Kim and P.~Panda, ``{Optimizing Deeper Spiking Neural Networks for Dynamic Vision Sensing},'' \emph{Neural Networks}, vol. 144, pp. 686--698, 2021.

\bibitem{li2022event}
Z.~Li, M.~S. Asif, and Z.~Ma, ``Event transformer,'' \emph{arXiv preprint arXiv:2204.05172}, 2022.

\bibitem{wu2023stca}
X.~Wu, Y.~Song, Y.~Zhou, Y.~Jiang, Y.~Bai, X.~Li, and X.~Yang, ``Stca-snn: self-attention-based temporal-channel joint attention for spiking neural networks,'' \emph{Frontiers in Neuroscience}, vol.~17, 2023.

\bibitem{hao2023reducing}
Z.~Hao, T.~Bu, J.~Ding, T.~Huang, and Z.~Yu, ``Reducing ann-snn conversion error through residual membrane potential,'' in \emph{Proceedings of the AAAI Conference on Artificial Intelligence (AAAI)}, 2023.

\bibitem{guo2022recdis}
Y.~Guo, X.~Tong, Y.~Chen, L.~Zhang, X.~Liu, Z.~Ma, and X.~Huang, ``Recdis-snn: rectifying membrane potential distribution for directly training spiking neural networks,'' in \emph{Proceedings of the IEEE/CVF Conference on Computer Vision and Pattern Recognition (CVPR)}, 2022.

\bibitem{zhang2019spike}
W.~Zhang and P.~Li, ``{Spike-Train Level Backpropagation for Training Deep Recurrent Spiking Neural Networks},'' in \emph{Advances in Neural Information Processing Systems (NeurIPS)}, vol.~32, 2019.

\bibitem{cheng2020lisnn}
X.~Cheng, Y.~Hao, J.~Xu, and B.~Xu, ``{LISNN: Improving Spiking Neural Networks with Lateral Interactions for Robust Object Recognition},'' in \emph{International Joint Conference on Artificial Intelligence (IJCAI)}, 2020, pp. 1519--1525.

\bibitem{loshchilov2017decoupled}
I.~Loshchilov and F.~Hutter, ``Decoupled weight decay regularization,'' \emph{arXiv preprint arXiv:1711.05101}, 2017.

\bibitem{kamata2022fully}
H.~Kamata, Y.~Mukuta, and T.~Harada, ``Fully spiking variational autoencoder,'' in \emph{Proceedings of the AAAI Conference on Artificial Intelligence}, vol.~36, no.~6, 2022, pp. 7059--7067.

\bibitem{TAID}
\BIBentryALTinterwordspacing
X.~Qiu, Z.~Luan, Z.~Wang, and R.~Zhu, ``When spiking neural networks meet temporal attention image decoding and adaptive spiking neuron,'' in \emph{The First Tiny Papers Track at {ICLR} 2023, Tiny Papers @ {ICLR} 2023, Kigali, Rwanda, May 5, 2023}, K.~Maughan, R.~Liu, and T.~F. Burns, Eds.\hskip 1em plus 0.5em minus 0.4em\relax OpenReview.net, 2023. [Online]. Available: \url{https://openreview.net/pdf?id=MuOFB0LQKcy}
\BIBentrySTDinterwordspacing

\bibitem{zhan2023esvae}
Q.~Zhan, X.~Xie, G.~Liu, and M.~Zhang, ``Esvae: An efficient spiking variational autoencoder with reparameterizable poisson spiking sampling,'' \emph{arXiv preprint arXiv:2310.14839}, 2023.

\bibitem{horowitz20141}
M.~Horowitz, ``1.1 computing's energy problem (and what we can do about it),'' in \emph{2014 IEEE International Solid-State Circuits Conference Digest of Technical Papers (ISSCC)}.\hskip 1em plus 0.5em minus 0.4em\relax IEEE, 2014, pp. 10--14.

\end{thebibliography}
\bibliographystyle{IEEEtran}
\end{document}